\documentclass{article} 
\usepackage{iclr2026_conference,times}
\usepackage[ruled,vlined]{algorithm2e}

\usepackage{amsmath,amsfonts,bm}

\def\eqref#1{equation~\ref{#1}}

\def\1{\bm{1}}

\DeclareMathAlphabet{\mathsfit}{\encodingdefault}{\sfdefault}{m}{sl}
\SetMathAlphabet{\mathsfit}{bold}{\encodingdefault}{\sfdefault}{bx}{n}

\usepackage{listings}
\usepackage[dvipsnames]{xcolor}
\usepackage{tikz}
\usetikzlibrary{positioning,calc,decorations.pathreplacing}
\usepackage{url}
\usepackage{xcolor}
\usepackage{booktabs}
\usepackage{wrapfig}
\usepackage{amssymb}
\usepackage{color}
\usepackage{colortbl}
\usepackage[table]{xcolor}
\usepackage{caption}
\usepackage{minitoc}
\makeatletter
\AtBeginDocument{%
  \addtocontents{toc}{\protect\setcounter{tocdepth}{-10}}
}
\makeatother

\usepackage[colorlinks]{hyperref}  
\hypersetup{
citecolor=[HTML]{3366CC}
}

\hypersetup{
linkcolor=[HTML]{3366CC}
}
\definecolor{mmada_color}{HTML}{ECECEC}
\definecolor{mmada_color}{HTML}{EFF7FF}
\newcommand{\method}{\textbf{MMaDA-Parallel}\xspace}

\title{MMaDA-Parallel: Multimodal Large Diffusion Language Models for Thinking-Aware Editing and Generation}

\author{Ye Tian$^{1,2* }$ \quad Ling Yang$^{3}\thanks{Equal Contribution.} $ \quad Jiongfan Yang$^{1}$ \quad Anran Wang$^{2}$\quad  Yu Tian$^{2}$ \quad Jiani Zheng$^{2}$  \\ \textbf{Haochen Wang $^{2,4}$} \quad \textbf{Zhiyang Teng}$^{2}$ \quad  \textbf{Zhuochen Wang$^{2}$}  \quad \textbf{Yinjie Wang$^{5}$} \quad  \\ \textbf{Yunhai Tong$^{1}\thanks{Correponding Authors}$} \quad \textbf{Mengdi Wang$^{3\dagger}$} \quad \textbf{{Xiangtai Li$^{2}$}}  \\
$^{1}$Peking University $^{2}$ByteDance $^{3}$Princeton University $^{4}$CASIA $^{5}$The University of Chicago  \\
Huggingface: \href{https://github.com/tyfeld/MMaDA-Parallel}{MMaDA-Parallel-Model} \quad Code: \href{https://github.com/tyfeld/MMaDA-Parallel}{MMaDA-Parallel-Code} 
}

\iclrfinalcopy 
\begin{document}

\maketitle

\begin{abstract}

While thinking-aware generation aims to improve performance on complex tasks, we identify a critical failure mode where existing sequential, autoregressive approaches can paradoxically degrade performance due to error propagation. 
To systematically analyze this issue, we propose ParaBench, a new benchmark designed to evaluate both text and image output modalities. Our analysis using ParaBench reveals that this performance degradation is strongly correlated with poor alignment between the generated reasoning and the final image.
To resolve this, we propose a parallel multimodal diffusion framework, \textbf{MMaDA-Parallel}, that enables continuous, bidirectional interaction between text and images throughout the entire denoising trajectory. MMaDA-Parallel is trained with supervised finetuning and then further optimized by Parallel Reinforcement Learning (ParaRL), a novel strategy that applies semantic rewards along the trajectory to enforce cross-modal consistency. Experiments validate that our model significantly improves cross-modal alignment and semantic consistency, achieving a 6.9\% improvement in Output Alignment on ParaBench compared to the state-of-the-art model, Bagel, establishing a more robust paradigm for thinking-aware image synthesis.
\end{abstract}



\section{Introduction}
\label{sec:intro}

Recent advances in multimodal generative models have achieved remarkable progress in instruction-based image generation and editing~\citep{esser2024scaling, flux, wei2024omniedit, liu2025step1xeditpracticalframeworkgeneral}. 
Given diverse textual prompts, these models can produce visually coherent and semantically aligned results across a wide range of tasks. 
However, these models often struggle with \textbf{complex instructions that require reasoning over world knowledge}, frequently leading to incorrect editing and generation~\citep{wu2025kris, niu2025wise, zhao2025envisioning}. 
To mitigate this gap, recent studies have introduced intermediate reasoning steps before visual generation~\citep{fang2025got, jiang2025t2i, deng2025emerging}. 
In these approaches, textual reasoning is first performed to guide subsequent image editing and generation. 
Such explicit reasoning has proven effective in improving the quality and consistency of image editing and generation~\citep{deng2025emerging}.

\begin{figure}[ht]
    \centering
    \includegraphics[width=\linewidth]{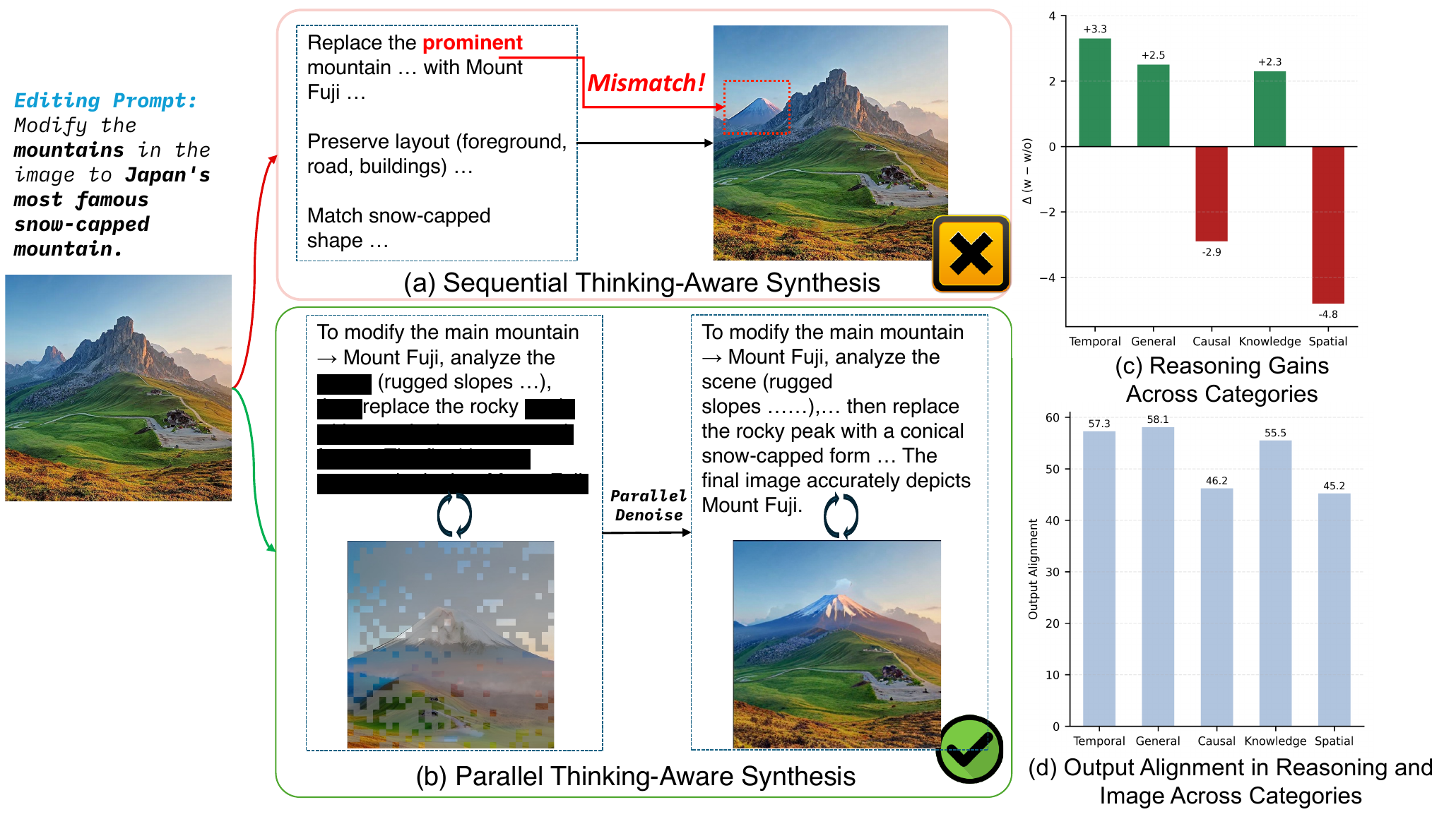}
    \caption{Sequential vs. parallel thinking-aware image synthesis. (a) Sequential generation (\textbf{Bagel, GPT4o}) may suffer from vague or incorrect reasoning. (b) Parallel generation aligns text and image at each denoising step, reducing hallucination and errors. (c) Quantitative comparison shows reasoning can degrade performance in certain categories. (d) Poorer categories also exhibit weaker reasoning–image alignment, highlighting the need for stronger cross-modal alignment.}
    \vspace{-0.26in}
    \label{fig:teaser}
\end{figure}

Despite the general effectiveness of incorporating a reasoning process prior to image synthesis, we observe a counterintuitive and critical phenomenon. On certain benchmarks~\citep{wu2025kris}, the inclusion of reasoning can in fact \textbf{reduce the semantic fidelity of the generated images}. For example, in Figure~\ref{fig:teaser}(a), a "thinking-aware" model starts with correct reasoning but then shifts to refining minor details like background textures. This reduces attention on the primary subject and causes the final edit to misidentify it completely. The resulting image thus deviates from the user's core instruction and even contradicts its own thinking prompt, leading to a clear performance drop.
%
%
This raises a crucial question: \textit{What underlies this performance degradation?}

Based on these failure cases, we hypothesize that the degradation stems from the reasoning text itself. However, this hypothesis is difficult to verify with existing benchmarks~\citep{wu2025kris, zhao2025envisioning}. These benchmarks only evaluate the final image against the initial prompt, but cannot evaluate the intermediate reasoning step or its alignment with the final output. 

Therefore, we introduce \textit{ParaBench}, our new benchmark designed to explicitly evaluate this output alignment between a model's generated reasoning and its final image. Using ParaBench to evaluate the state-of-the-art model Bagel~\citep{deng2025emerging}, we find a strong correlation: performance degradation occurs precisely in categories where output alignment is weakest (Figure~\ref{fig:teaser}(d)). 
We attribute this to the compounding errors inherent in sequential autoregressive models, where ambiguous or incomplete reasoning provides unreliable guidance for the subsequent image generation, ultimately degrading the final output.

Thus, while pre-reasoning can in principle enhance multimodal generation, its reliance on an autoregressive pipeline makes the process vulnerable to error accumulation and semantic drift. 
Recently, another line of work has explored discrete diffusion models for text or image generation~\citep{nie2025large, yang2025mmada, dream2025}, which remove the token-by-token constraint of autoregression and instead employ confidence-based sampling to achieve greater global consistency. 
Inspired by these advances, we ask: \textbf{What if multimodal models could generate text and images in parallel?} Such a paradigm directly addresses the limitations of AR reasoning: text and images can attend to each other at every denoising step, avoiding the propagation of hallucinations and vague priors while grounding textual descriptions in visual evidence.

Building on this insight, we propose a purely diffusion-based framework for \textit{parallel text–image generation}, where cross-modal interaction is maintained throughout the trajectory to ensure robust and semantically faithful multimodal editing and generation, as shown in Figure~\ref{fig:teaser}(b)).

\begin{figure}[t]
    \centering
    \includegraphics[width=\linewidth]{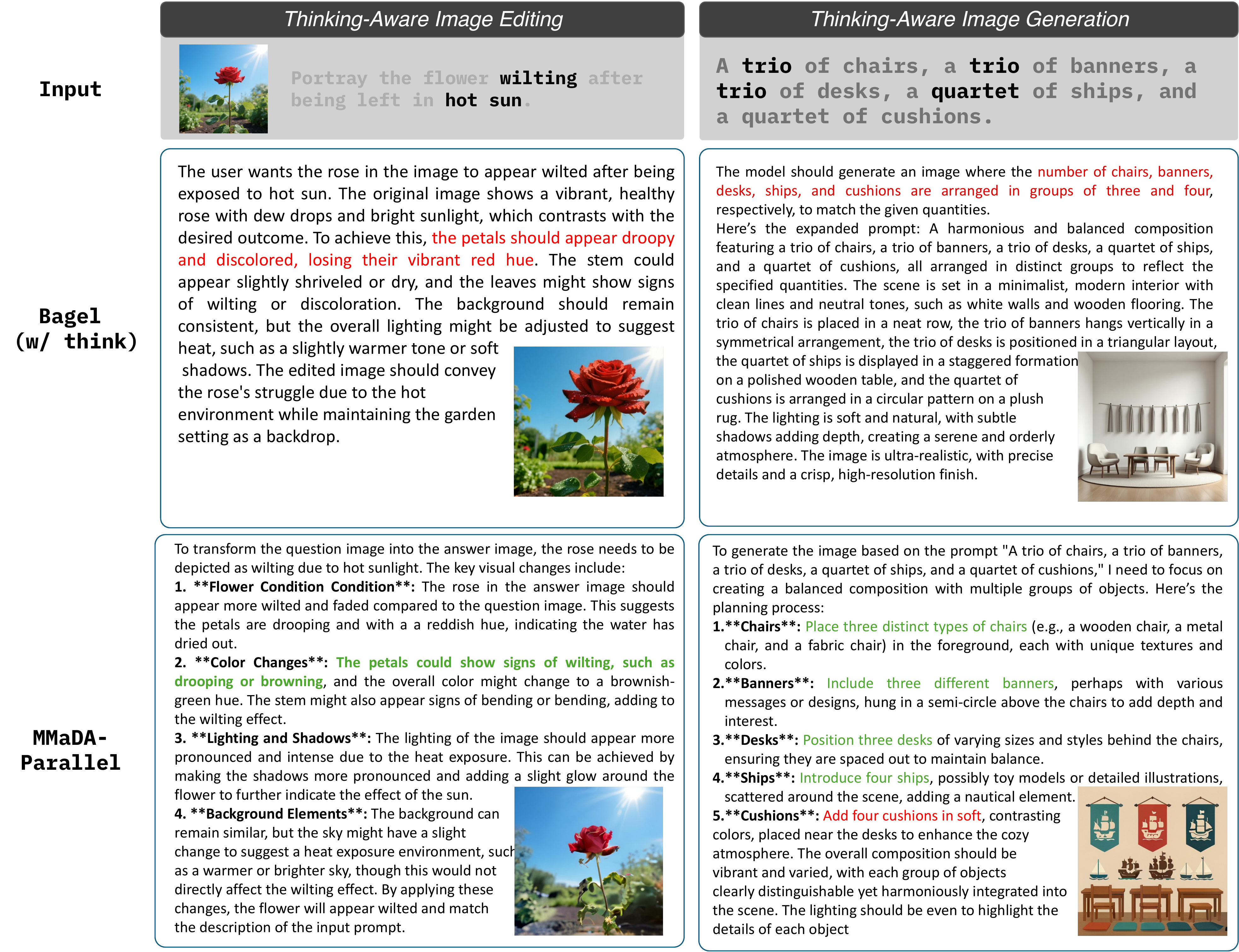}
    \caption{\method supports parallel, thinking-aware image editing and generation. Compared with Bagel, \method demonstrates superior reasoning quality and stronger alignment between the generated text and image outputs.}
    \vspace{-0.2in}
    \label{fig:teaser2}
\end{figure}
To train this framework, we first establish a thinking-aware data curation pipeline. We prompt a powerful VLM with data triplets ($\langle \text{input image}, \text{edit instruction}, \text{output image} \rangle$) sourced from widely-adopted image editing and generation datasets. The VLM is tasked to generate a reasoning trace that explains the edit process.
This pipeline yields a training dataset of quadruplets: $\langle \text{input image}, \text{instruction}, \text{reasoning trace}, \text{output image} \rangle$, designed to elicit the models' reasoning and generation capabilities.
We use this dataset to perform supervised fine-tuning on MMaDA~\citep{yang2025mmada}.
This parallel version, \method, demonstrates higher output consistency compared to sequential baselines, as can be observed in Figure~\ref{fig:teaser2}.

Notably, such consistency is observed not only in the final outputs but also \textbf{throughout the generation trajectory.} We observe that during the parallel denoising process, the image region corresponding to a specific semantic concept is often refined simultaneously with its textual counterpart. 
However, standard SFT and conventional reinforcement learning algorithms optimize for the final outcome only. This output-level supervision is too coarse to enforce the fine-grained, stepwise alignment we observe and cannot guarantee consistency at intermediate steps. 
To fully leverage this trajectory-level consistency, we draw inspiration from process-level and trajectory-level optimization methods~\citep{li2024process, wang2025revolutionizing}  and introduce \textit{Parallel Reinforcement Learning (ParaRL)}. %
Instead of focusing solely on the final outcome, ParaRL incorporates stepwise semantic supervision to refine alignment along the denoising trajectory. Our experiments demonstrate that this trajectory-level optimization provides a more granular and effective signal for diffusion models compared to traditional output-level supervision.

Extensive quantitative and qualitative results validate the effectiveness of MMaDA-Parallel for thinking-aware image editing and generation, and further highlight the additional gains achieved through ParaRL. Our contributions can be summarized as follows:

\begin{enumerate}
    \item \textbf{In-depth Benchmarking and Analysis of Thinking-aware Image Synthesis.} We propose ParaBench, which systematically evaluates thinking-aware image generation and editing, focusing on text and image quality and their alignment.
    \item  \textbf{Parallel Multimodal Diffusion Framework.} We propose a purely discrete diffusion-based approach for parallel thinking-aware image editing and generation, which enables bidirectional attention between modalities at every denoising step and effectively alleviates the error accumulation of autoregressive pipelines.
    \item \textbf{Parallel Reinforcement Learning.} We introduce a parallel reinforcement learning strategy, \textit{ParaRL}, which assigns semantic rewards along the denoising trajectory, further enhancing alignment between the output modalities and the overall performance.
    \item \textbf{Extensive Evaluation and State-of-the-Art Alignment.} Our comprehensive experiments validate the framework, establishing state-of-the-art performance among open-source models with a 6.9\% gain in Output Alignment over Bagel on our ParaBench benchmark, while maintaining comparable performance on single-modality metrics.
\end{enumerate}


\section{Related Work}
\label{sec:related_work}


Recent progress in multimodal models for image understanding, generation, and editing has been rapid, yet most approaches remain constrained to single-modal generation conditioned on multiple modalities~\citep{sd3, wu2025qwen, labs2025flux1kontextflowmatching, bai2025qwen2}. To improve the accuracy and fidelity of multimodal generation, a growing line of work has explored introducing a textual \textit{Chain-of-Thought}  reasoning process before image generation or editing. We refer to this paradigm as \textbf{thinking-aware image generation and editing}. For instance, early efforts such as Chameleon~\citep{team2024chameleon} and {Mogao}~\citep{liao2025mogao} investigated interleaved generation, enabling interleaving sequences of text and image tokens. Image-CoT~\citep{guo2025can} and GoT~\citep{fang2025got} incorporated CoT reasoning before image synthesis, demonstrating that reasoning traces can enhance generation quality. {Bagel}~\citep{deng2025emerging} further extended this idea by integrating chain-of-thought reasoning into both image generation and editing, enabling more flexible and semantically aligned outputs. Building on this direction, follow-up works such as {OmniGen2}~\citep{wu2025omnigen2} and {IRG}~\citep{huang2025interleaving} introduced reflective reasoning after image generation, using multi-turn textual feedback to refine visual outputs iteratively.
Most existing methods, however, rely on a sequential autoregressive interleaved pipeline, which could limit direct cross-modal interaction and make the model prone to error accumulation from imperfect reasoning traces. Exploring a parallel generation framework that enables more interaction within output modalities is still lacking in this scenario.
(More related work can be found in Appendix~\ref{app:more_related}).

\section{MMaDA-Parallel}
\label{sec:method}

\subsection{Findings and Benchmarking on Thinking-Aware Synthesis}
\label{bench}

To investigate whether pre-generation reasoning genuinely enhances performance, we conduct a controlled study on image editing tasks, which provides a clearer instruction-grounded evaluation than naive synthesis. 
We sample inputs from established benchmarks~\citep{wu2025kris, zhao2025envisioning} and generate paired outputs using Bagel~\citep{deng2025emerging}—an advanced, open-source, unified model supporting thinking-aware generation—with and without thinking. 
We report the average editing evaluation metrics in Kris-Bench~\citep{wu2025kris} in Figure~\ref{fig:teaser}(c) and also Table~\ref{tab:krisbench-extended}.



\textbf{Findings.} While the reasoning step enhanced performance on most tasks, a notable countertrend emerged: performance declined in a significant subset of cases, about 23\%, particularly in complex compositional edits. 
A closer analysis reveals that these failures often stemmed from low-quality or vague reasoning text, which misguides the image generation process. 
This exposes a critical gap in existing protocols: they evaluate the final image but ignore the quality of the intermediate reasoning—the other generated modality.

\textbf{Benchmarking mixed modalities.}  This analysis reveals a fundamental limitation in current evaluation paradigms: existing benchmarks~\citep{wu2025kris, zhao2025envisioning, ghosh2023geneval} only evaluate images, ignoring the quality of the reasoning itself and its consistency with the image. 
To address this gap, we introduce \textbf{ParaBench}, a new benchmark specifically designed for the comprehensive evaluation of thinking-aware image synthesis. 
ParaBench comprises 300 challenging prompts, split into 200 for editing and 100 for generation. The editing prompts are meticulously curated to test a wide spectrum of abilities, covering not only general operations (e.g., add, remove, replace) but also complex tasks requiring reasoning. 
The 100 generation prompts focus on open-ended creative synthesis of complex scenes. 
We evaluate models on ParaBench using the GPT-4.1 across six fine-grained aspects: for the textual output, we assess Text Quality and Text Alignment; for the visual output, we evaluate Image Quality, Image Alignment, and Image Consistency; and finally, the overall Output Alignment between them. 
More details are included in Appendix~\ref{sec:appendix_parabench}.

\begin{table}[t]
\centering
\caption{\textbf{Thinking may degrade the performace of visual synthesis.} Bagels' performance comparison on ParaBench editing tasks with and without thinking. We also report the reasoning quality (Text Qual.) and cross-modal alignment (Output Align.).}
\label{tab:krisbench-extended}
\resizebox{0.9\linewidth}{!}{
\begin{tabular}{l|ccccc}
\toprule
Editing Category & w/o Thinking & w/ Thinking & $\Delta$ (w/ $-$ w/o)& Text Qual. $\uparrow$  & Output Align.$\uparrow$ \\
\midrule
 Temporal  & 72.3 & 75.6 & +3.3& 92.6 & 57.3 \\
General    & 68.9 & 71.4 & +2.5 & 86.2 & 58.1\\
 Causal & 70.1 & 67.2 & \textbf{$-$2.9} & \textbf{75.3} & \textbf{46.2}\\
Knowledge  & 74.5 & 76.8  & +2.3& 87.8 & 55.5\\
 Spatial   & 69.8 & 65.0  & \textbf{$-$4.8} & \textbf{73.2} & \textbf{45.2}\\
\bottomrule
\end{tabular}}
\end{table}


To demonstrate ParaBench's diagnostic capabilities, we apply it to a representative baseline, Bagel. While full quantitative results are presented in Sec~\ref{sec:bench}, Table~\ref{tab:krisbench-extended} highlights a crucial finding by focusing on two key metrics: \textbf{Text Quality} and \textbf{Output Alignment}. The results reveal a clear correlation between the quality of the reasoning step and the final performance. 
Notably, the categories that exhibited performance degradation also suffered from significant drops in both reasoning quality and reasoning-image synergy. This pattern strongly suggests that poor reasoning does not merely fail to provide helpful guidance but actively misleads the generation process, validating the necessity of explicitly improving the synergy between text and image generation.

\textbf{Motivations on parallel multimodal diffusion.} Our benchmarking results reveal a critical limitation in current thinking-aware generation: \textit{the sequential generation paradigm}, where reasoning precedes image synthesis, creates a rigid dependency that can propagate errors and limit cross-modal synergy. 
When reasoning quality degrades, it directly undermines the subsequent image generation, as demonstrated by the correlated performance drops in spatial and temporal editing tasks. 
To address this fundamental issue, we propose a parallel unified multimodal diffusion framework that enables simultaneous generation of both reasoning text and images, fostering genuine multimodal collaboration while eliminating the error propagation inherent in sequential approaches.

\begin{figure}[h]
    \centering
    \includegraphics[width=1\linewidth]{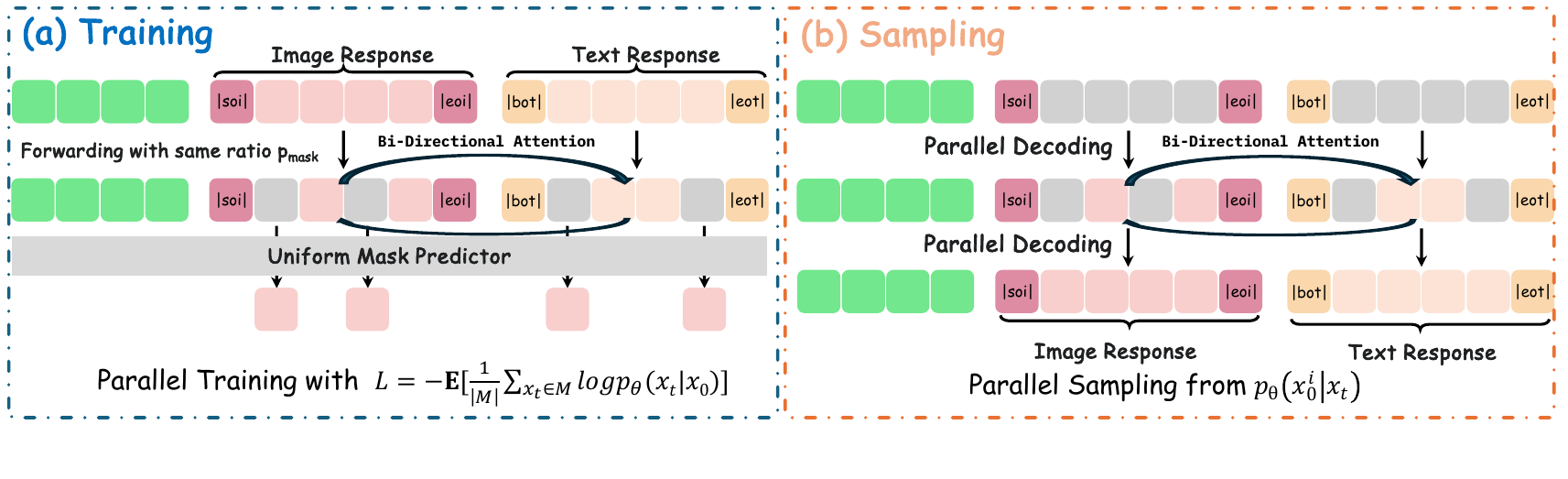}
    \caption{Parallel Generation Architecture: During (a) training, image and text responses are masked and predicted in parallel with a uniform mask predictor, optimized by the masked token likelihood objective. During (b) sampling, the model performs parallel decoding to generate both image and text responses jointly, enabling efficient multimodal response generation.}
    \label{fig:parallel}
\end{figure}

\subsection{Basic Algorithm and Architecture}
\label{sec:algo}

Discrete diffusion models have demonstrated strong performance for both image and text generation~\citep{bai2024meissonic, nie2025large, llada1.5}. 
Building on the unified discrete-diffusion view, MMaDA~\citep{yang2025mmada} demonstrates that a single diffusion framework can jointly model multiple modalities; however, its decoding remains \textit{sequential} across modalities. 
To overcome this limitation, we propose a \textit{parallel} multimodal diffusion framework that: (i) represents all modalities as discrete tokens, (ii) arranges them in an interleaved sequence with bidirectional attention, and (iii) employs a single mask predictor shared across modalities, enabling synchronous denoising for both text and images. 
An overview of this framework is shown in Figure~\ref{fig:parallel}. 


\paragraph{Interleaved discrete sequence layout.}
Following the MMaDA framework~\citep{yang2025mmada}, we process both text and images within a unified discrete token space. 
Specifically, we tokenize text using the LLaDA tokenizer~\citep{nie2025large} and encode images into a grid of discrete visual tokens using a pretrained MAGVIT-v2~\citep{magvitv2} quantizer. 
These tokenized modalities are then serialized into a single interleaved sequence, using explicit sentinels and task tags to enable full bidirectional cross-modal attention:

\begin{verbatim}
    Input:  <|task|><|soi|>[img]<|eoi|><|bos|>[text]<|eos|>
    Output: <|soi|>[output img]<|eoi|><|bos|>[output text]<|eos|>
\end{verbatim}
During training, we concatenate the input and output templates into a single sequence, allowing the model to attend from outputs to inputs within a unified context. 
The task token \texttt{<|task|>} is instantiated differently depending on the scenario, with \texttt{<|thinkgen|>} used for thinking-aware generation and \texttt{<|thinkedit|>} used for thinking-aware editing. 
This single-sequence design eliminates the ordering asymmetry and exposure bias introduced by autoregressive cross-modal pipelines.

\paragraph{Training objective.}
Let $x_0\in \{1,\ldots,V\}^{L}$ denote the concatenated training sequence (input part followed by output part), where $L$ is the total number of tokens in the sequence. We keep the input part static and apply noise only to the output part. 
At a sampled timestep $t \in \{1,\ldots,T\}$, for each token in the \textit{output} part we replace it with \texttt{[MASK]} with probability $\beta_t$ and keep it unchanged with probability $1-\beta_t$; tokens in the \textit{input} part are left unchanged:
\begin{equation}
x_t^{(i)}=
\begin{cases}
x_0^{(i)} & \text{if $i$ in input},\\[2pt]
x_0^{(i)} \ \text{with prob. } (1-\beta_t),\ \texttt{[MASK]} \ \text{with prob. } \beta_t & \text{if $i$ in output.}
\end{cases}
\label{eq:noising_step}
\end{equation}
Equivalently, for positions in the output, the absorbing-state marginal after $t$ steps is $q(x_t \mid x_0)=\alpha_t\,x_0 + (1-\alpha_t)\,\mathbf{m}$ where $\alpha_t=\prod_{k=1}^{t}(1-\beta_k),$ and $\mathbf{m}$ is the one-hot distribution of  \texttt{[MASK]}. 

The parallel diffusion model $p_\theta(\cdot \mid x_t)$ is formulated as a unified masked-token predictor over the joint vocabulary of text and image tokens. 
Let $i \in {1,\ldots,L}$ denote token positions in the concatenated input–output sequence. 
Since only the output segment is noised during diffusion, the model predicts ground-truth tokens $x_0$ at the currently masked positions within this segment. 
To better balance the training dynamics across modalities, we make the timestep-dependent loss weight modality-specific: tokens in the \textit{output image} segment and the \textit{output text} segment are assigned separate weights, $w_{\text{img}}(t)$ and $w_{\text{text}}(t)$. 
For compactness, we write the objective using a unified token-aware weight function $w(t,i)$.
We optimize a timestep-reweighted cross-entropy:
\begin{equation}
\mathcal{L}_{\text{parallel}}(\theta)
= - \,\mathbb{E}_{t,\,x_0,\,x_t}
\Bigg[
\sum_{i=1}^{L} 
w(t,i)\,\mathbf{1}\big[x_t^{(i)}=\texttt{[MASK]}\big]\,
\log p_\theta\!\big(x_0^{(i)} \mid x_t\big)
\Bigg],
\label{eq:unified_loss_modality}
\end{equation}
where $\mathbf{1}[\cdot]$ is the indicator function and 
\[
w(t,i)=
\begin{cases}
w_{\text{img}}(t), & \text{if $i$ lies in the \textit{output image} segment},\\[4pt]
w_{\text{text}}(t), & \text{if $i$ lies in the \textit{output text} segment}.
\end{cases}
\]
We empirically find that applying a timestep-dependent weighting $w_{\text{text}}(t)=1/t$ for text tokens and a constant weighting $w_{\text{img}}(t)=1$ for image tokens substantially stabilizes the training of image quality and output alignment. 
We illustrate this process in Figure~\ref{fig:parallel}(a) and include detailed additional preliminaries with ablations in Appendix~\ref{app:pre}.

\paragraph{Parallel denoising with dual schedulers.}
Decoding proceeds along a shared diffusion time axis $t_T \!\rightarrow\! \cdots \!\rightarrow\! t_0$, as is shown in Figure~\ref{fig:parallel}(b). We define two modality-specific schedulers,
$u_{\text{img}}(t), u_{\text{text}}(t) \in [0,1]$, which specify the target proportion of unmasked tokens at step $t$.
At each reverse step:
(i) the model jointly predicts distributions for all currently masked positions;
(ii) for each modality, a fraction of tokens is sampled (e.g., via confidence-based sampling), while the remaining positions are retained as \texttt{[MASK]}.
Because attention is bidirectional across the \textit{entire} sequence, text and image can inform each other at every step of decoding. In our experiments, the text schedule is implemented as a fully linear reveal schedule combined with semi-autoregressive confidence-based decoding~\cite{nie2025large}, while the image schedule follows a cosine reveal schedule with global confidence-based decoding. More details can be found in Appendix~\ref{app:sampling}.

\subsection{Post Training with Parallel Reinforcement Learning}

\begin{figure}[t]
    \centering
    \includegraphics[width=1\linewidth]{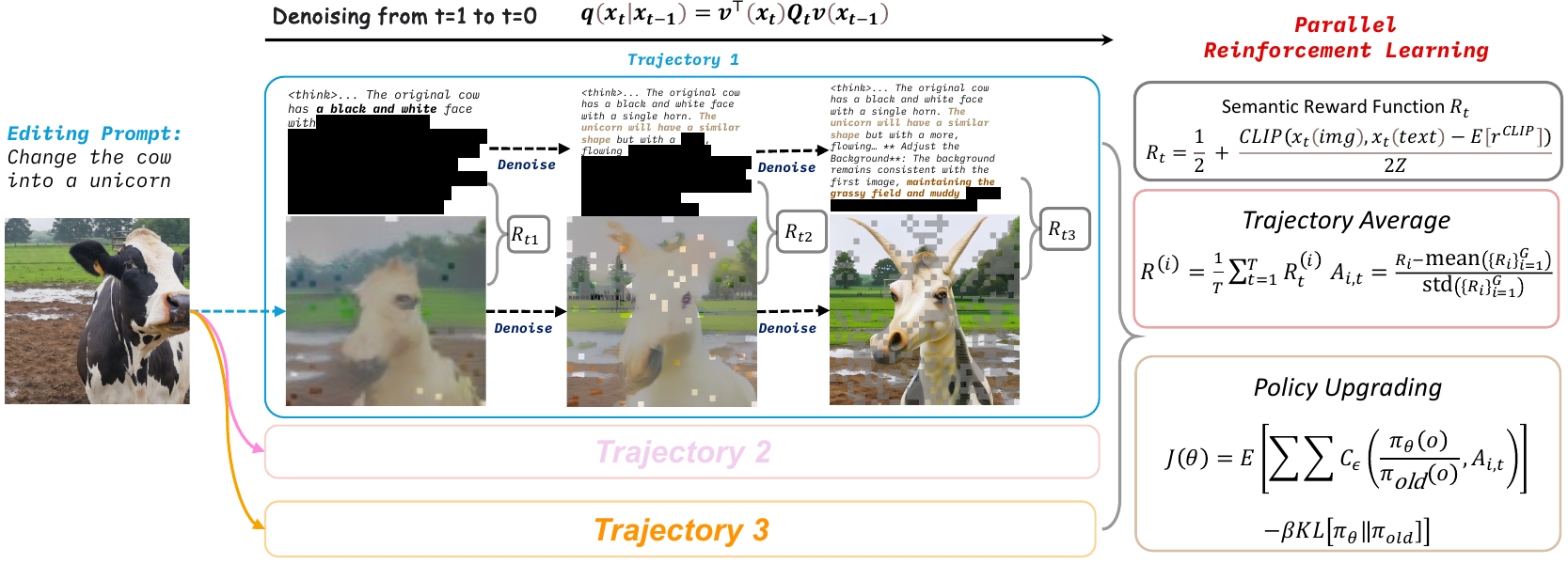}
    \caption{Overview of our proposed Parallel Reinforcement Learning (ParaRL). Rather than  optimization only to the final denoised outputs, ParaRL introduces reward signals along the entire denoising trajectory, reinforcing semantic alignment consistently throughout the generation process.}
    \label{fig:tspo}
    \vspace{-1.5em}
\end{figure}

\paragraph{Supervised Finetuning for Parallel Synthesis}
A key challenge in our approach is that existing generation and editing datasets lack the reasoning traces required for our parallel synthesis framework. 
To address this, we construct a suitable training dataset by first aggregating samples from various sources. 
For each sample comprising an input image (for editing tasks), an instruction, and the final output image, we employ a multimodal LLM (Qwen-2.5-VL in our implementation) to generate a corresponding reasoning trace. 
Further details on the dataset construction process, including the sources and categories, are provided in Appendix~\ref{sec:appendix_dataset}. 
We then use this dataset to perform supervised fine-tuning on MMaDA~\citep{yang2025mmada}.
This process adapts it into a parallel variant capable of performing thinking-aware synthesis, where reasoning and generation occur concurrently. 


\begin{wrapfigure}{r}{0.45\textwidth}
  \begin{center}
    \includegraphics[width=0.45\textwidth]{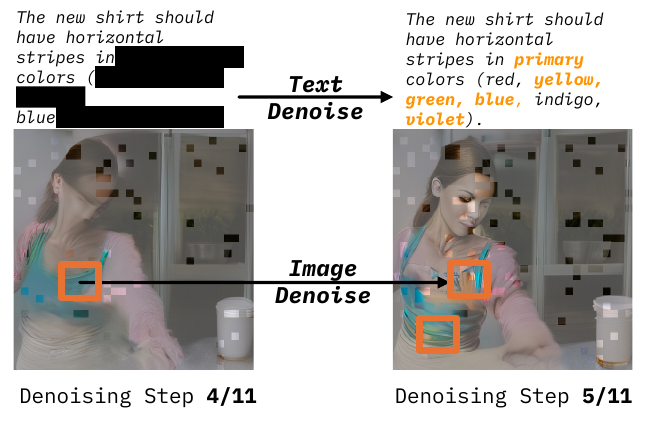}
  \end{center}
    \caption{Synergy of sampling. Given the prompt: ``change the blue shirt to a vibrant rainbow color,'' the specific color decoding in text and image emerges at the same step.}
  \vspace{-1em}
  \label{fig:synergy}
\end{wrapfigure}
\paragraph{Synergy along the denoising trajectory.} While analyzing generations from the finetuned model, we observe that certain semantic concepts emerge \textit{synchronously} in text and image at intermediate denoising steps. 
As illustrated in Figure~\ref{fig:synergy}, when tasked to change a shirt to a "vibrant rainbow color," the specific color words and their corresponding visual features appear at the same timestep. 
This observation leads to a key insight: cross-modal alignment is not an endpoint phenomenon but is progressively established \textbf{throughout the generation trajectory}. 
This implies that supervision applied to these intermediate steps, not just the final output, can further improve this alignment.

\paragraph{Parallel reinforcement learning with trajectory optimization.}
Building on this insight, we further introduce Parallel Reinforcement Learning (ParaRL), a novel training paradigm that directly leverages this intermediate cross-modal synergy. 
Instead of rewarding only the final output, ParaRL uses the alignment between text and image tokens at each denoising step as a dense reward signal. 

Specifically, for a given query $Q$, the generated response is a full trajectory $\tau_i \triangleq \big(\tau_i(1),\dots,\tau_i(|\tau_i|)\big)$, where $|\tau_i|$ is the total number of denoising steps and $\tau_i(t)$ is the set of tokens decoded at step $t$. 
While this formulation provides a step-wise reward $r_{i,t}$ for each intermediate response $\tau_i(t)$, optimizing over the entire dense trajectory is computationally prohibitive. 
To make training feasible, we adopt a sparse optimization strategy. During each online rollout, we pre-select sampling steps $s$ and fix subset of step indices $S \subset \{1, \dots, |\tau_i|\}, |S|=s$ and only compute rewards $r_{i,t}$ and their corresponding standardized advantages $A_{i,t}$ for timesteps $t \in S$.
We adapt a diffusion GRPO objective~\citep{gong2025diffucoder} that accommodates token-level likelihood ratios with advantages calculated at these sampled steps:
\begin{equation}
\begin{aligned}
\mathcal{J}_{\text{policy}}(\theta)
= & \; \mathbb{E}_{\substack{Q \sim D_{\text{task}} \\ \{\tau_i\}_{i=1}^G \sim \pi_{\text{old}}(\cdot \mid Q)}}
\left[
\sum_{i=1}^G \sum_{t \in S}
\frac{1}{|\tau_i(t)|}
\sum_{o \in \tau_i(t)}
C_\epsilon\!\left(
\frac{\pi_{\theta}(o \mid Q, \tau_i(1{:}t-1))}
     {\pi_{\text{old}}(o \mid Q, \tau_i(1{:}t-1))},\,
A_{i,t}
\right)
\right] \\
& - \beta \, \mathrm{KL}\!\big[\pi_{\theta}\,\|\,\pi_{\text{old}}\big],
\end{aligned}
\label{eq:objpolicy}
\end{equation}
where $C_\epsilon(r, A)\triangleq \min\!\big(rA,\, \mathrm{clip}(r,\,1-\epsilon,\,1+\epsilon)\,A\big)$. In this objective, the summation is performed over the sparsely sampled steps $t \in S$. 
The term $o$ ranges over all tokens within the state $\tau_i(t)$ at a sampled step $t$, and $\tau_i(1{:}t-1)$ denotes the full history of tokens generated prior to step $t$. 
Finally, $\pi_{\text{old}}$ is the behavior policy for generating rollouts, and $\beta$ controls the KL penalty strength.
\paragraph{Trajectory reward design.}
In typical trajectory-level optimization frameworks, a well-trained process reward model (PRM)~\citep{li2024process} or value function~\cite{wang2025revolutionizing} is often required, since intermediate partial outputs usually lack sufficient semantic information for reliable evaluation. 
Surprisingly, in our parallel text–image generation setting, we find that intermediate fragments are already semantically meaningful.
For instance, even partially decoded text tokens often reveal enough semantic cues to compute alignment with the simultaneously generated image content, as illustrated in Figure~\ref{fig:tspo}. 
This observation allows us to bypass the need for a dedicated PRM: we directly employ \textit{semantic alignment} between text and image as the reward signal. 

Unlike tasks with binary rewards (e.g., mathematical reasoning), our cross-modal alignment objective provides a continuous reward signal. 
However, the naive CLIP score, which serves as our reward source, can exhibit high variance and an arbitrary scale, making it unstable for direct use in reinforcement learning. 
To ensure training stability, we therefore apply a normalization scheme inspired by prior work in RL with continuous rewards~\citep{liu2025flow}. 
We begin by estimating the mean $\mu_{\text{CLIP}}$ and standard deviation $\sigma_{\text{CLIP}}$ of CLIP scores across the training distribution, where we compute on a random 1\% subset of the data.
Let $c_{i,t} = R^{\text{CLIP}}(\text{text}(\tau_i(t)), \text{image}(\tau_i(t)))$ be the raw CLIP score for the content generated at step $t$. We first standardize this score to obtain $\hat{c}_{i,t}$ using $\hat{c}_{i,t} = \frac{c_{i,t} - \mu_{\text{CLIP}}}{\sigma_{\text{CLIP}}}$. 
This standardized score is then clipped to the range $[-1, 1]$ and linearly rescaled to yield the final reward $R_{i,t}$, which is bounded within $[0, 1]$:
\begin{equation}
R_{i,t} = \frac{1}{2} \left( 1 + \mathrm{clip}(\hat{c}_{i,t}, -1, 1) \right)
\label{eq:reward}
\end{equation}
The corresponding advantages $A_{i,k}$ used in Eq.~\ref{eq:objpolicy} are then obtained by standardization over the rollouts:
$A_{i,t} = \frac{R_{i,t} - \mathrm{mean}\!\left(\{R_{j,t}\}_{j=1}^G\right)}
                      {\mathrm{std}\!\left(\{R_{j,t}\}_{j=1}^G\right)} $



\section{Experiments}

\subsection{Implementation Details}

\paragraph{Training and datasets.}
Our final model, MMaDA-Parallel, is trained in a two-stage process. 
We begin with supervised finetuning (SFT) on the MMaDA-MixCoT model, which integrates a LLaDA-8B text backbone with a MagVIT-v2 image tokenizer. 
For this stage, we construct a new dataset of 150K thinking-aware image editing and generation pairs, meticulously sourced and filtered from multiple existing benchmarks. 
In the second stage, we apply reinforcement learning with a GRPO-based objective. To enhance training efficiency, this RL stage focuses on the most challenging 10\% of the SFT examples, optimizing the policy online to improve cross-modal semantic alignment. 
More details of the dataset and training details can be found in Appendix~\ref{sec:appendix_dataset} and ~\ref{sec:appendix_implementation}.

\paragraph{Evaluation setup.}
We conduct our primary evaluation on the ParaBench benchmark, which was introduced in the Method section. We employ an LLM-as-a-judge framework (GPT-4.1) to assess performance across the six fine-grained metrics previously described, covering text quality, image fidelity, and cross-modal alignment. The prompts used for the LLM judge are detailed in the Appendix~\ref{sec:appendix_parabench}. Our \method is compared against state-of-the-art thinking-aware models, including Bagel~\citep{deng2025emerging}, GPT-4o, and Gemini-2.5, as well as leading image-only generators like Qwen-Image~\citep{wu2025qwen}, Qwen-Image-Edit~\citep{wu2025qwen}, Flux.1-dev~\citep{flux} and Flux.1-Kontext~\citep{labs2025flux1kontextflowmatching}.

\subsection{Main Results}
\label{sec:main_results}
\begin{figure}[t]
    \centering
    \vspace{-1em}
    \includegraphics[width=0.98\linewidth]{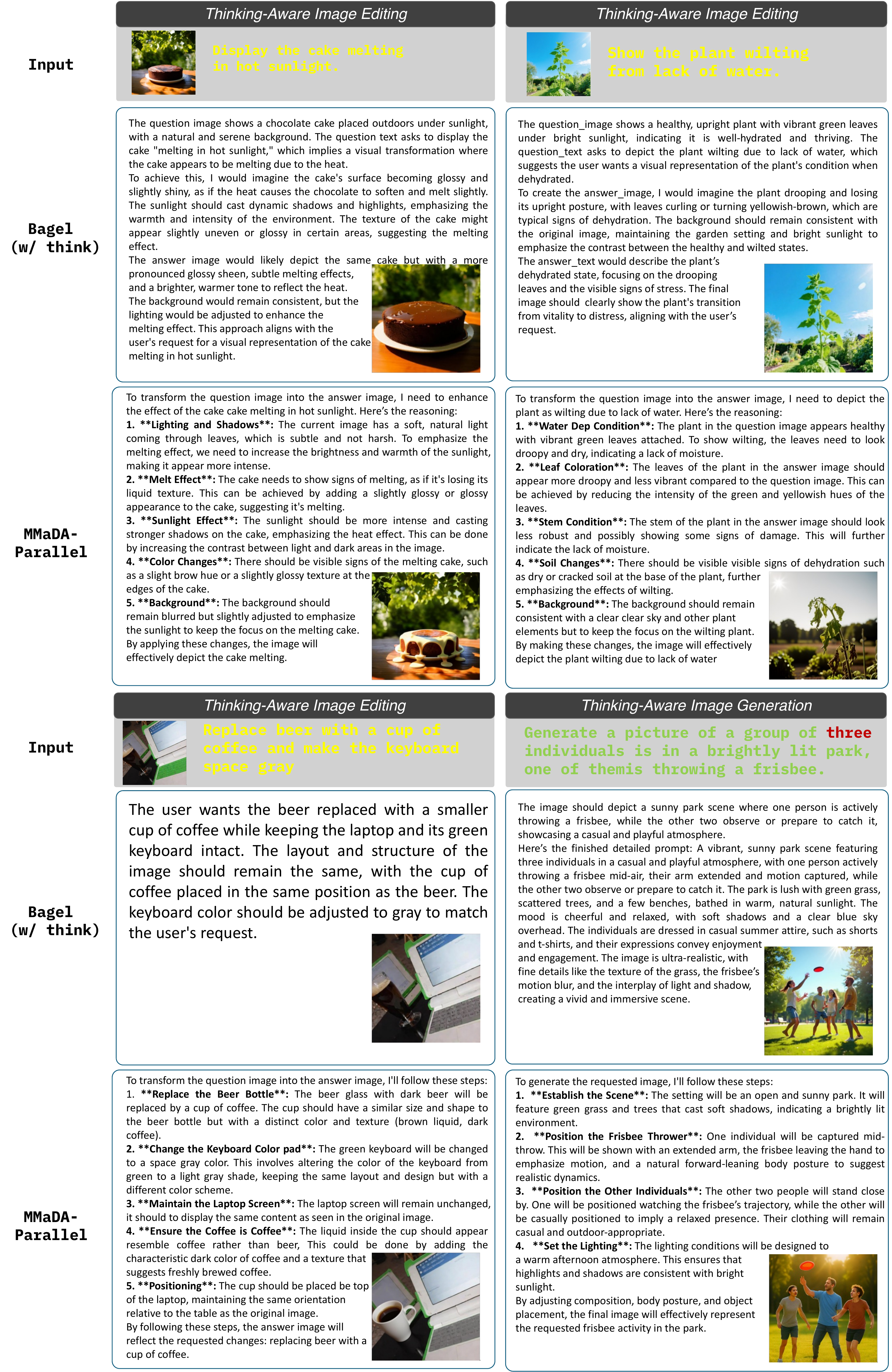}
    \caption{Qualitative results in comparison with Bagel.}
    \label{fig:Qualitative Results}
    \vspace{-2.em}
\end{figure}

\begin{table*}[t]
\centering
\caption{\textbf{Main results on \emph{ParaBench}}. 
Evaluation across all editing and generation tasks. For non-thinking image editing or generation models, text evaluation and output alignment cannot be computed. }
\label{tab:main_results}
\resizebox{\textwidth}{!}{
\begin{tabular}{l|cccccc|c}
\toprule
\textbf{Model} & \textbf{Text Qual.} & \textbf{Text Align.} & \textbf{Image Cons.} & \textbf{Image Align.} & \textbf{Image Qual.} & \textbf{Output Align.} & \textbf{Overall} \\
\midrule

\multicolumn{8}{l}{\textbf{Open-source models (Non-thinking)}} \\
\midrule
Flux.1-Dev                                               & -                             & -                             & -                             & 65.2                          & 77.5                          & -                                           &        -          \\
Qwen-Image                                                                                        & -                             & -                             & -                             & \underline{67.2}                    & \textbf{84.2}                 & -                                           &     -             \\
Flux.1-Kontext                                                                                    & -                             & -                             & \underline{77.9}                    & 65                            & 84                            & -                                           &      -            \\
Qwen-Image-Edit                                                                                   & -                             & -                             & \textbf{78.2}                 & \textbf{73.5}                 & {84.1}                    & -                                           &        -          \\
Bagel (w/o think)                                                                                 & -                             & -                             & 72.2                          & 50.3                          & 80.1                          & -                                           &       -           \\

\midrule
\multicolumn{8}{l}{\textbf{Closed-source models}} \\
\midrule
GPT-4o                                                                                            & 92.5                          & 93.4                          & 86.2                          & \textbf{85.7}                 & 88.1                          & \textbf{69.5}                               & \textbf{85.9}    \\
Gemini-2.5                                                                                        & \textbf{94.1}                 & \textbf{95.2}                 & \textbf{88.5}                 & 76.2                          & \textbf{90.2}                 & 63.4                                        & 84.6             \\

\midrule
\multicolumn{8}{l}{\textbf{Open-source models (Thinking-aware)}} \\
\midrule
Bagel (w/ think)                                                                                  & \textbf{82}                   & 70.5                          & \textbf{76.7}                 & \textbf{63.4}                 & \textbf{81.5}                 & \underline{52.9}                                  & {71.2}       \\
Show-o* (tuned)                                                                                   & 75.2                          & \underline{70.7}                    & 69.1                          & 57.5                          & 78.5                          & 48.9                                        & 66.6             \\

\midrule
\rowcolor{mmada_color}\textbf{\method w/o Para-RL}      & 76.5                          & 70.4                          & 70.5                          & 58.2                          & 80.5                          & 51.5                                        & 67.9             \\
\rowcolor{mmada_color}\textbf{\method  w/ Para-RL}       & \underline{80.4}                    & \textbf{71}                   & \underline{73.4}                    & \textbf{63.2}                    & \textbf{81.2}                    & \textbf{59.8}                               & \textbf{71.5} \\
\bottomrule
\end{tabular}}
\end{table*}


Table~\ref{tab:main_results} reports the overall performance on our ParaBench benchmark. Our proposed method, \method, achieves the highest \textit{Output Alignment} among all open-source models, confirming the effectiveness of its parallel multimodal decoding and trajectory-level optimization. In terms of general text and image quality, \method performs on par with Bagel, despite Bagel being trained on a dataset nearly three orders of magnitude larger. Compared to leading closed-source models like GPT-4o and Gemini-2.5, \method substantially narrows the gap in alignment metrics while maintaining competitive text and image quality, demonstrating remarkable data efficiency. Furthermore, the results indicate that our ParaRL stage consistently improves output text-image consistency, suggesting that trajectory-level optimization effectively strengthens cross-modal grounding throughout the generation process.

In addition, we provide a qualitative comparison with open-source models in Figure~\ref{fig:Qualitative Results}, showcasing examples of both editing and generation. A key observation is that \method produces more precise and descriptive reasoning traces. This enhanced reasoning leads to superior visual fidelity in the final image. For instance, our model accurately renders complex instructions like a "melting cake" and correctly applies causal reasoning to depict "withered grass." Moreover, \method demonstrates stronger compositional abilities, particularly in counting, correctly generating "three people" or "two faces of a clock" where Bagel often fails. In contrast, Bagel's reasoning in these challenging cases tends to be vague or omits crucial details, leading to inaccurate image synthesis. These results further underscore \method's capability for advanced thinking-aware editing and generation, driven by better-aligned semantic information.


\subsection{Analysis of Key Contributions}
\begin{table*}[ht]
\centering
\begin{minipage}{0.48\textwidth}
\centering
\caption{Parallel vs sequential decoding.}
\vspace{-1em}
\label{tab:rq1}
\resizebox{\textwidth}{!}{
\begin{tabular}{l|ccc}
\toprule
\textbf{Denoising} & \textbf{Text Align.} & \textbf{Image Align.} & \textbf{Output Align.} \\
\midrule
Sequential &   70.6   &   56.1   &  48.9    \\
\textbf{Parallel}       &  \textbf{70.4}    & \textbf{58.2}     &  \textbf{51.5}    \\
\bottomrule
\end{tabular}}
\end{minipage}
\hfill
\begin{minipage}{0.48\textwidth}
\centering
\caption{Output vs trajectory-level RL.}
\vspace{-1em}
\label{tab:rq2}
\resizebox{\textwidth}{!}{
\begin{tabular}{l|ccc}
\toprule
\textbf{Model} & \textbf{Text Align.} & \textbf{Image Align.} & \textbf{Output Align.} \\
\midrule
before RL  &  70.4    & 58.2     &  51.5    \\
w/ Output-level RL &   70.7   &   62.3   &  53.6    \\
\textbf{w/ ParaRL (Ours)}     &   \textbf{71}   &   \textbf{63.2}   &  \textbf{59.8}    \\
\bottomrule
\end{tabular}}
\end{minipage}
\end{table*}

\begin{table*}[ht]
\centering
\caption{Ablation on sampling steps $s$ in ParaRL.}
\vspace{-1em}
\label{tab:abl_chunk}
\resizebox{\textwidth}{!}{
\begin{tabular}{l|cccccc|c}
\toprule
\textbf{ParaRL $s$} & \textbf{Text Qual.} & \textbf{Text Align.} & \textbf{Image Cons.} & \textbf{Image Align.} & \textbf{Image Qual.} & \textbf{Output Align.} & \textbf{Overall} \\
\midrule
Before RL & 76.5                          & 70.4                          & 70.5                          & 58.2                          & 80.5                          & 51.5                                        & 67.9             \\
ParaRL $s{=}2$  & 77.9 & 70.3 & 71.5 & 62.8 & 80.7 & 53.6 & 68.6 \\
\rowcolor{mmada_color} ParaRL ($s{=}3$) \,\,\textit{(default)}  & \underline{80.4}                    & \textbf{71.0}                   & \textbf{73.4}                    & \underline{63.2}                    & \textbf{81.2}                    & \textbf{59.8}                               & \textbf{71.5} \\
ParaRL ($s{=}4$)  & \textbf{80.5}  & 70.8 & \underline{73.2} & \textbf{63.5} & \underline{80.8} & \underline{58.7} & \underline{71.3}  \\
\bottomrule
\end{tabular}}
\end{table*}
After presenting the overall results, we now return to the two central research questions that motivated our work:
\textbf{RQ1:} Does parallel denoising improve generation quality compared with sequential denoising?
\textbf{RQ2:} Does trajectory-level finetuning improve over output-level finetuning?

\paragraph{The Benefit of Parallel Decoding (RQ1).}
We compare our model against a sequential baseline (\emph{MMaA-Sequential}) that generates text before images. During training, noise was applied to only one modality at a time to align with this sequential inference process.  Table~\ref{tab:rq1} shows our parallel framework substantially outperforms this baseline on key alignment metrics, with comparable text and image quality. This result validates our core hypothesis: simultaneous, interactive decoding is crucial for reducing error propagation and producing coherent multimodal outputs.

\begin{figure}[ht]
    \centering
    \begin{minipage}{0.48\linewidth}
        \centering
        \includegraphics[width=\linewidth]{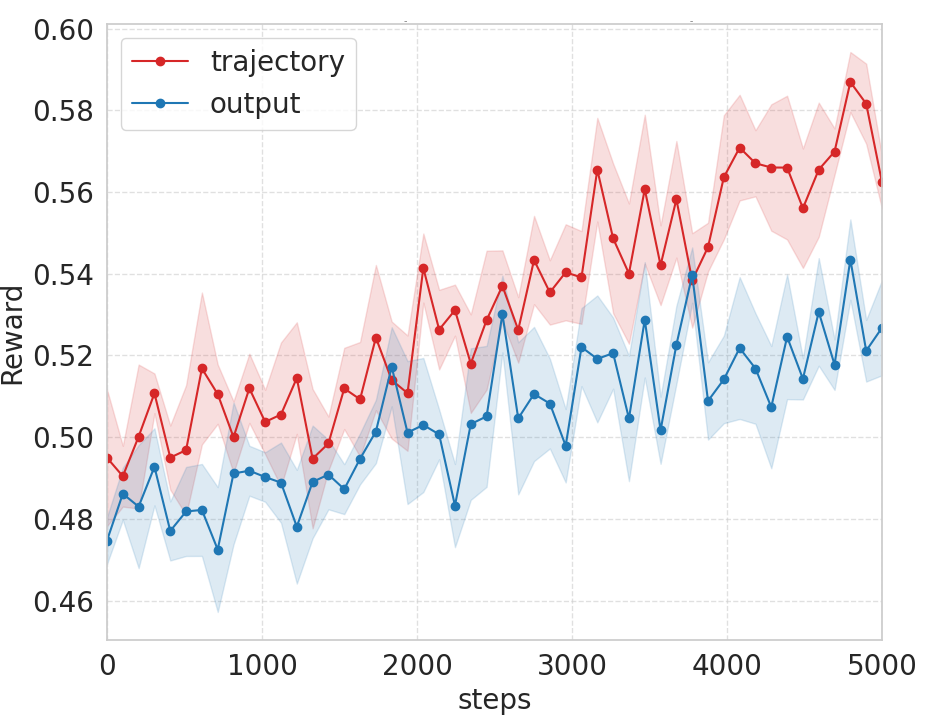}
         \vspace{-1.5em}
        \caption{ParaRL reward training curve between trajectory and output level optimization.}
        \label{fig:sub1}
    \end{minipage}
    \hfill
    \begin{minipage}{0.48\linewidth}
        \centering
        \includegraphics[width=\linewidth]{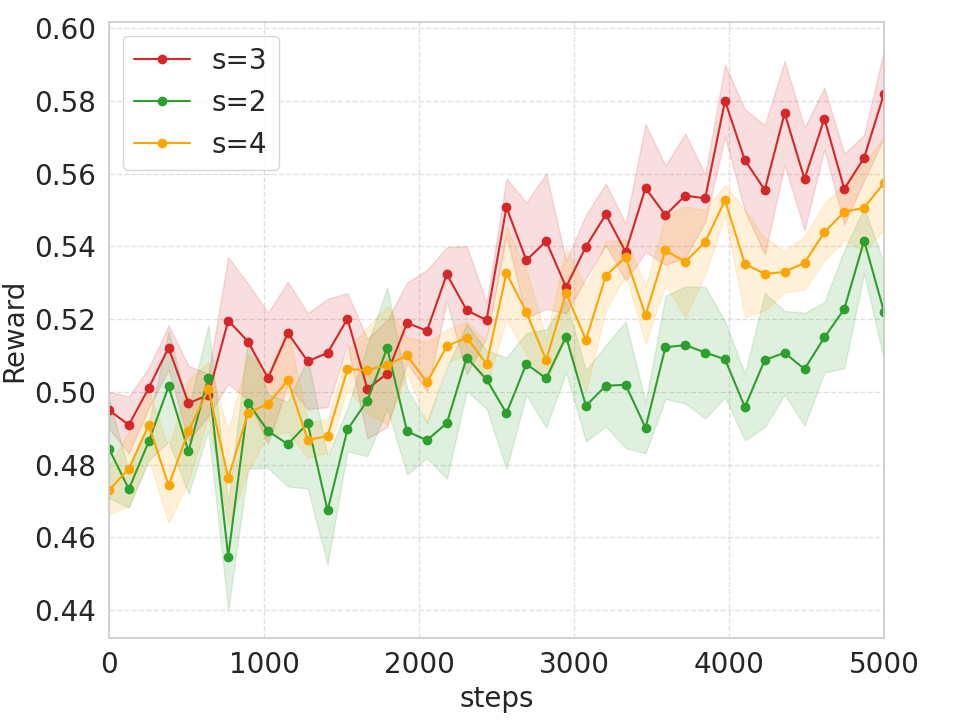}
         \vspace{-1.5em}
        \caption{ParaRL reward training curve across different sampling steps of the trajectory.}
        \label{fig:sub2}
    \end{minipage}
\end{figure}

\paragraph{The Benefit of Trajectory-Level Optimization (RQ2).}
We compare two reinforcement learning strategies:  (i) \emph{output-level RL}, where rewards are computed on the final generated sample, and (ii) our proposed \emph{ParaRL} with trajectory-level finetuning, where rewards are aggregated across denoising steps.  As shown in Table~\ref{tab:rq2}, trajectory-level optimization yields gains in text–image consistency and output alignment, and Figure~\ref{fig:sub1} further shows that it enables more stable training dynamics.

Another key hyperparameter in this strategy is the number of sampled steps, $s$. We analyze its impact in Table~\ref{tab:abl_chunk} and report the training curve in Figure~\ref{fig:sub2}. We find that using $s=3$ or $s=4$ yields substantial improvements over $s=2$, as a denser reward signal provides more stable guidance. We adopt $s=3$ in the final configuration for the best balance between performance and efficiency.

\section{Conclusion}
In this work, we investigated a critical phenomenon where sequential thinking-aware models can paradoxically suffer from performance degradation on complex tasks. We conducted an in-depth analysis using our proposed ParaBench benchmark, which uniquely evaluates both output modalities, and found a strong correlation between this degradation and poor alignment between the generated modalities. To resolve this, we propose a parallel multimodal diffusion framework trained with supervised finetuning and further optimized by Parallel Reinforcement Learning (ParaRL)—our novel method of applying rewards along the entire denoising trajectory. Experiments validate that our approach significantly improves cross-modal alignment and semantic consistency, establishing a more robust paradigm for thinking-aware image synthesis.



\newpage
\appendix
\addtocontents{toc}{\protect\setcounter{tocdepth}{2}} 

\begingroup
\hypersetup{linktoc=none} 
\renewcommand{\contentsname}{Appendix Contents}
\tableofcontents
\endgroup







\section{Scaling of \method}
\begin{figure}[ht]
    \centering
    \includegraphics[width=\linewidth]{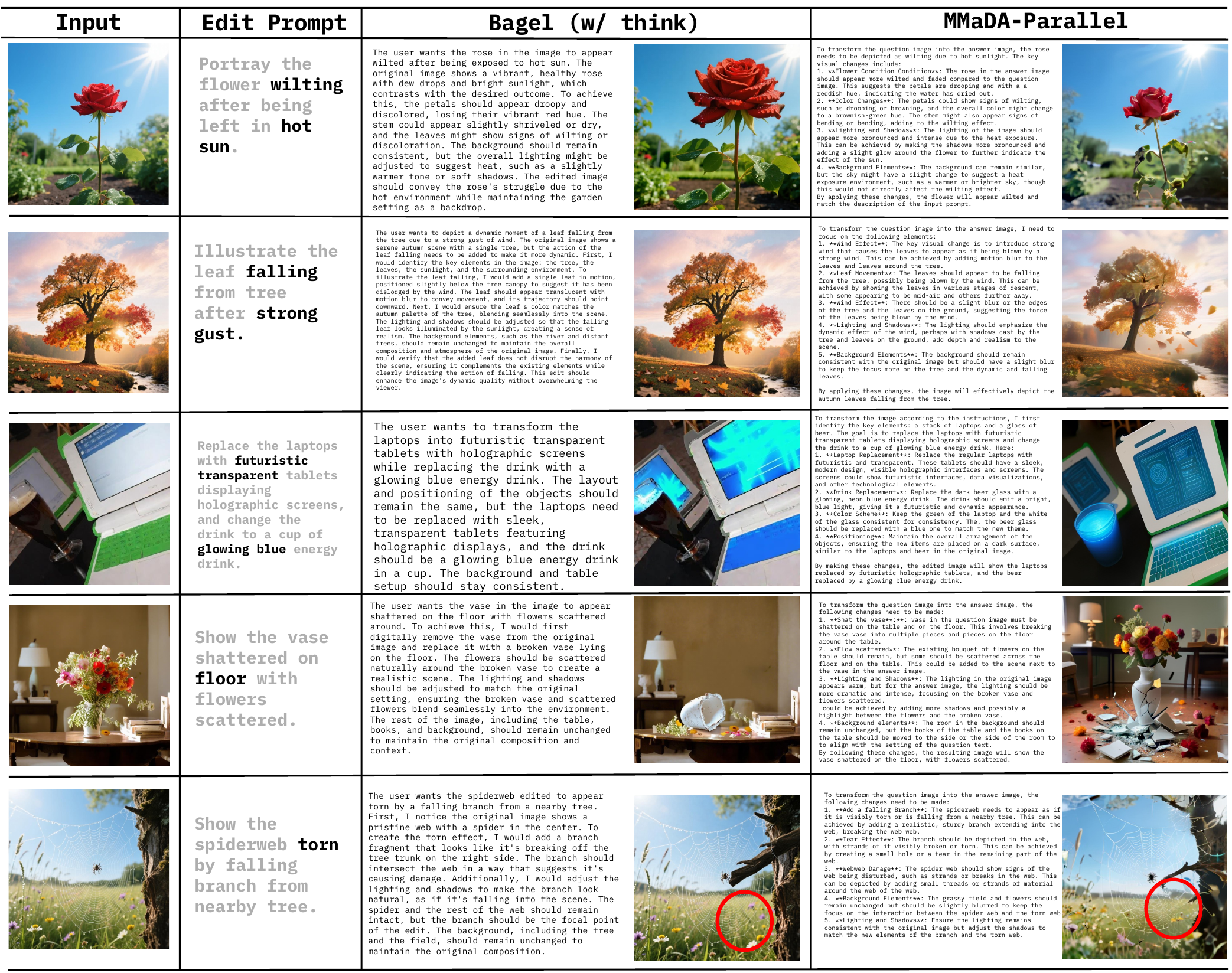}
    \caption{Additional qualitative results with the same training settings.}
    \label{fig:lumina}
\end{figure}

To further validate our \method on larger-scale training, we extend our post-training framework on Lumina-DiMOO~\cite{xin2025lumina}. Lumina-DiMOO shares a similar architecture with MMaDA, but benefits from much larger-scale data training. same training settings for Lumina-DiMOO, and report its corresponding quantitative and qualitative results in Table~\ref{tab:main_results_lumina} and Figure~\ref{fig:lumina}. The results clearly show that after applying our Parallel framework and ParaRL post-training, Lumina-DiMOO surpasses Bagel and achieves new state-of-the-art performance in thinking-aware synthesis. This finding strongly validates the scalability of our method.

\label{sec:bench}
\begin{table*}[ht]
\centering
\caption{\textbf{Main results on \textit{ParaBench}}. Post-training with Lumina-DiMOO.}
\label{tab:main_results_lumina}
\vspace{-1em}
\resizebox{\textwidth}{!}{
\begin{tabular}{l|cccccc|c}
\toprule
\textbf{Model} & \textbf{Text Qual.} & \textbf{Text Align.} & \textbf{Image Cons.} & \textbf{Image Align.} & \textbf{Image Qual.} & \textbf{Output Align.} & \textbf{Overall} \\
\midrule

\multicolumn{8}{l}{\textbf{Open-source models (Non-thinking)}} \\
\midrule
Flux.1-Dev                                               & -                             & -                             & -                             & 65.2                          & 77.5                          & -                                           &        -          \\
Qwen-Image                                                                                        & -                             & -                             & -                             & \underline{67.2}                    & \textbf{84.2}                 & -                                           &     -             \\
Flux.1-Kontext                                                                                    & -                             & -                             & \underline{77.9}                    & 65                            & 84                            & -                                           &      -            \\
Qwen-Image-Edit                                                                                   & -                             & -                             & \textbf{78.2}                 & \textbf{73.5}                 & {84.1}                    & -                                           &        -          \\
Bagel (w/o think)                                                                                 & -                             & -                             & 72.2                          & 50.3                          & 80.1                          & -                                           &       -           \\

\midrule
\multicolumn{8}{l}{\textbf{Closed-source models}} \\
\midrule
GPT-4o                                                                                            & 92.5                          & 93.4                          & 86.2                          & \textbf{85.7}                 & 88.1                          & \textbf{89.5}                               & \textbf{89.2}    \\
Gemini-2.5                                                                                        & \textbf{94.1}                 & \textbf{95.2}                 & \textbf{88.5}                 & 76.2                          & \textbf{90.2}                 & 83.4                                        & 88.9             \\

\midrule
\multicolumn{8}{l}{\textbf{Open-source models (Thinking-aware)}} \\
\midrule
Bagel (w/ think)                                                                                  & 82.0                   & \underline{74.5}                          & \textbf{76.7}                 & 63.4                & 81.5                 & 52.9                                  & {71.8}       \\
Show-o* (tuned)                                                                                   & 75.2                          & 70.7                    & 69.1                          & 57.5                          & 78.5                          & 48.9                                        & 66.6             \\

\midrule
\rowcolor{mmada_color}\textbf{\method* w/o Para-RL}      & \underline{82.6}                          & 73.7                          & 71.3                          & \underline{64.6}                          & \underline{82.6}                          & \underline{63.3}                                     & \underline{73.0}          \\
\rowcolor{mmada_color}\textbf{\method*  w/ Para-RL}       & \textbf{84.1}                    & \textbf{76.5}                   & 71.0                    & \textbf{67.8}                    & \textbf{83.6}                    & \textbf{68.8}                               & \textbf{75.3} \\
\bottomrule
\end{tabular}}
\vspace{-1em}
\end{table*}

\section{Additional  Results}
\subsection{Qualitative results}
\label{app:quan}
\begin{figure}[ht]
    \centering
    \includegraphics[width=\linewidth]{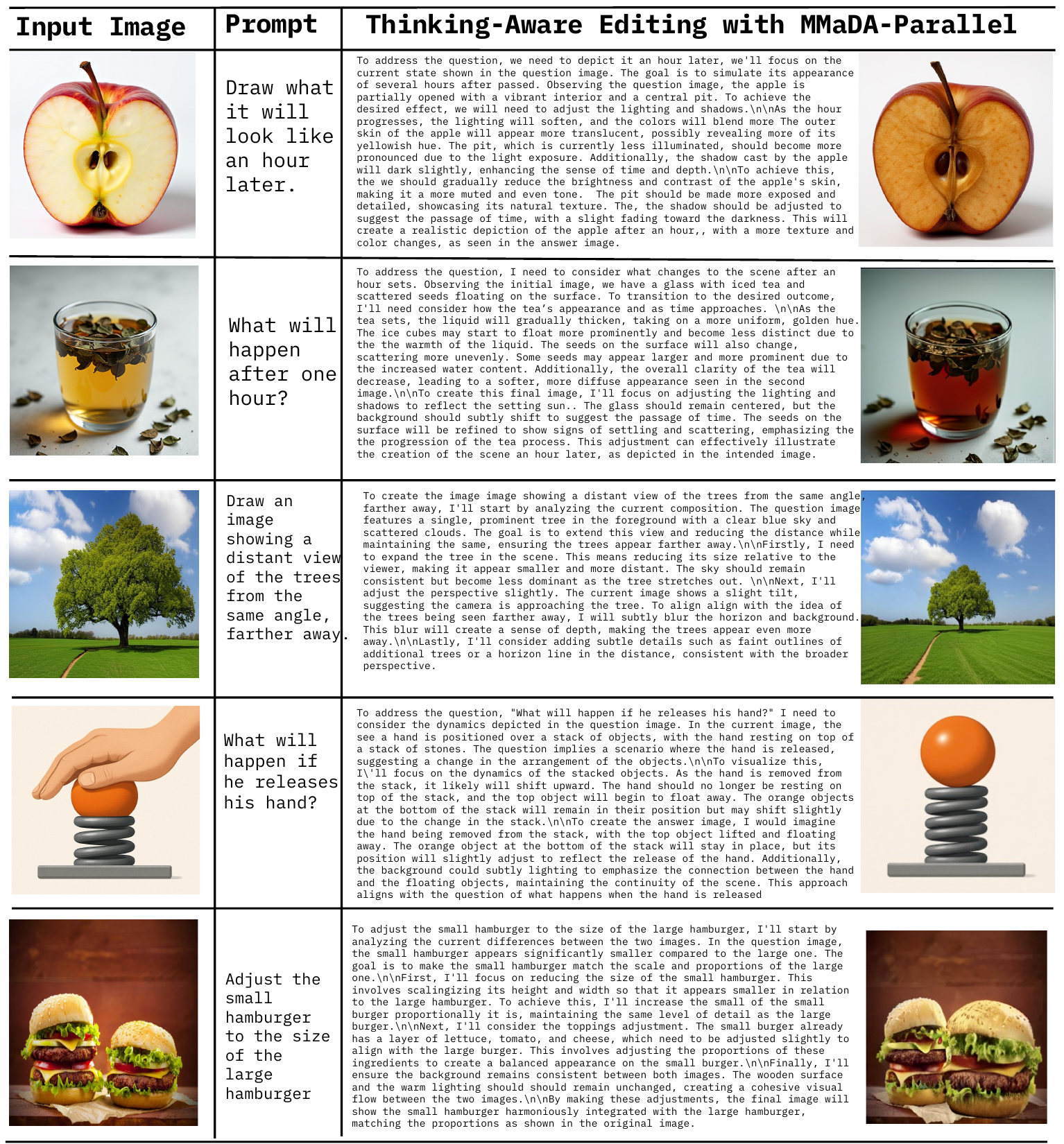}
    \caption{Additional qualitative results on thinking-aware image editing.}
    \label{fig:add_edit}
\end{figure}

\begin{figure}[ht]
    \centering
    \includegraphics[width=\linewidth]{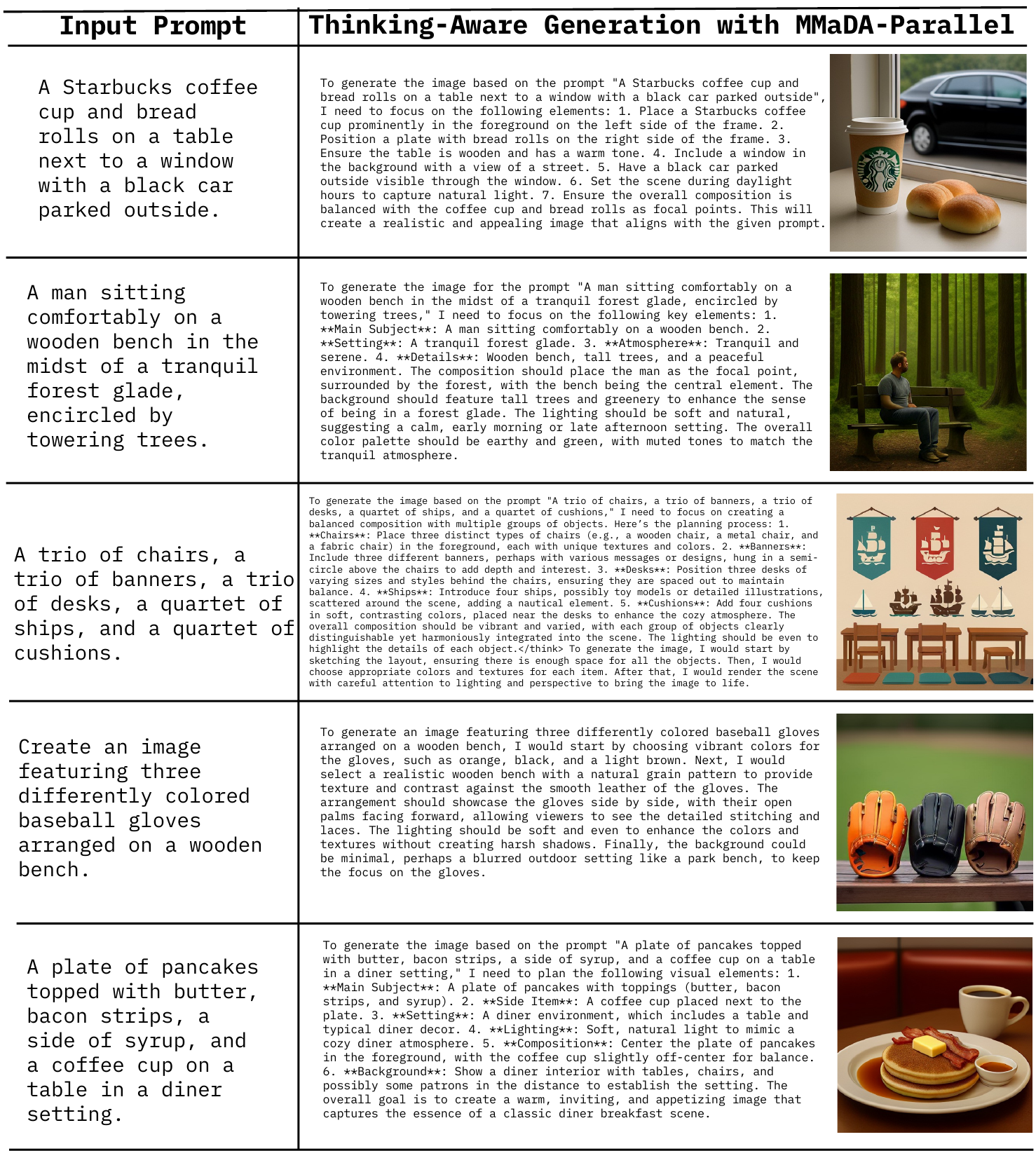}
    \caption{Additional qualitative results on thinking-aware image generation.}
    \label{fig:add_gen}
\end{figure}

We provide more qualitative results in Figure~\ref{fig:add_edit} and Figure~\ref{fig:add_gen} for thinking-aware image editing and generation.

\subsection{Quantitative Results}
We also report additional image-only results of \method on RISEBench~\cite{wu2025kris} and GenEval~\cite{ghosh2023geneval}. For fair comparison, we evaluate against the sequential version of MMaDA, where \method achieves consistent performance improvements, demonstrating that parallel generation leads to overall gains in image alignment. Compared with the original MMaDA, our approach further narrows the performance gap with Bagel.

\begin{table}[ht]\small
    \centering
    \caption{\textbf{Overall performance on RISEBench.} .}
    \label{rise}
    \vspace{1em}
    \resizebox{0.8\textwidth}{!}{%
    \begin{tabular}{l|cccc|c}
    \toprule
        \textbf{Models} & \textbf{Temporal} & \textbf{Causal} & \textbf{Spatial} & \textbf{Logical} & \textbf{Overall} \\
        \midrule
        GPT-4o-Image  & \textbf{34.1\%}   &\textbf{32.2\%}  & \textbf{37.0\%}  & \textbf{10.6\%} & \textbf{28.9\%} \\
        Gemini-2.0-Flash-exp  & 8.2\%   & 15.5\%   & 23.0\% & 4.7\% & 13.3\% \\
        BAGEL& 3.5\% & 4.4\% & 9.0\% & 5.9\% & 5.8\% \\
        \midrule
        MMaDA(Sequential) & 3.9 \% & 5.2\% & 8.1\% & 4.8\% & 5.5\%\\
        \rowcolor{mmada_color}MMaDA-Parallel & 4.2\% & 5.5\% & 8.3\% & 5.1\% & 5.75\%\\
        \bottomrule
    \end{tabular}}
    \label{tab:main1}
\end{table}

\begin{table*}[ht]
\centering
\caption{\textbf{Results on GenEval.}.
}
\resizebox{\linewidth}{!}{
\begin{tabular}{lccccccc}
\toprule
\textbf{Method} & \textbf{Single Obj.} & \textbf{Two Obj.} & \textbf{Counting} & \textbf{Colors} & \textbf{Position} & \textbf{Color Attri.} & \textbf{Overall} \\
\midrule
SDXL & \textbf{0.98} & 0.74 & 0.39 & \textbf{0.85} & 0.15 & 0.23 & 0.55 \\
Show-o~\cite{xie2024show} & 0.95 & 0.52 & 0.49 & 0.82 & 0.11 & 0.28 & 0.53 \\
MMaDA~\citep{yang2025mmada} & 0.99 & 0.76 & 0.61 & 0.84 & 0.20   & 0.37 & 0.63 \\
Bagel~\citep{deng2025emerging} & 0.98 & 0.95 &  0.84 & 0.95  & 0.78  & 0.77 &  0.88 \\
\midrule
MMaDA(Sequential) & 0.99 & 0.78 & 0.66 & 0.87 & 0.34   & 0.37 & 0.68\\
\rowcolor{mmada_color}MMaDA-Parallel & 0.99 & 0.83 & 0.70 & 0.88 & 0.40   & 0.47 & 0.71 \\
\bottomrule
\end{tabular}}

\label{tab:gen_eval_results}
\end{table*}

\section{More Related Work}
\label{app:more_related}
\paragraph{Diffusion large language models.}
Diffusion models have achieved remarkable progress in vision~\citep{ddpm,sd,sd3,ddim,dit}, motivating their extension to text. The discrete nature of textual tokens, however, makes direct adaptation non-trivial. Two main approaches have emerged: learning continuous latent representations~\citep{analog,tess,dinoiser,diffuseq}, and designing discrete diffusion models~\citep{yourdiscrete,scalingdiscrete,longllada,dream7b,llada1.5}. Among the latter, \textbf{Masked Diffusion Models (MDMs)} stand out by leveraging bidirectional attention for global consistency and supporting parallel decoding. Systems such as Dream7B~\citep{dream7b} and LLaDA~\citep{nie2025large} achieve performance comparable to autoregressive LLMs. 
Beyond text, diffusion-based LLMs have also been extended to multimodal domains. LaViDA~\citep{lavida} employs multi-view image encoding with masked-denoising training, LLaDA-V~\citep{lladav} integrates masked diffusion with visual instruction tuning, and MMaDA~\citep{yang2025mmada} unifies reasoning across text and vision generation through chain-of-thought supervision and reinforcement learning. These advances highlight the scalability and versatility of diffusion-based language models across both unimodal and multimodal settings. Nevertheless, existing approaches have not yet explored \textbf{parallel text–image co-generation}, leaving cross-modal reasoning and alignment still constrained by sequential pipelines.


\paragraph{Reinforcement learning for multimodal foundation models.}
Reinforcement Learning (RL) has emerged as a powerful paradigm for enhancing reasoning and controllability in large models. The widely adopted GRPO~\citep{guo2025deepseek} applies rewards primarily on the correctness of the final answer and the adherence to a predefined format. Recently, RL has been adopted in multimodal large language models~\citep{chen2025r1v,meng2025mm, r1-onevision,r1vl,openvlthinker, visionr1}, incorporating task-specific rewards such as answer correctness, intersection-over-union (IoU) for localization~\citep{liu2025seg}, and image–text alignment scores (e.g., T2I-R1~\citep{jiang2025t2i}). Extensions such as~\citep{jiang2025co,hong2025reinforcing} further introduce cross-modality coherence rewards. In the context of diffusion language models, similar strategies have been explored with verified rewards and carefully designed probability approximations~\citep{yang2025mmada,gong2025diffucoder}
.
Despite these advances, most existing methods focus solely on rewards applied to the final output, while largely ignoring the generative trajectory. 
This overlooks the fact that intermediate steps can provide crucial signals for alignment. In contrast, our work investigates the synergy between modalities during the denoising process and introduces ParaRL, which exploits stepwise semantic alignment to optimize thinking-aware multimodal generation.

\section{Preliminaries}
\label{app:pre}
\subsection{Preliminaries of discrete Diffusion Models.}

In recent years, diffusion models have set new standards in generative modeling. While Denoising Diffusion Probabilistic Models (DDPMs) excel in continuous domains like raw pixel spaces, Discrete Denoising Diffusion Probabilistic Models (D3PMs) have proven highly effective for discrete data, such as tokenized images and text. Models like VQ-Diffusion~\cite{vqdiffusion}, MaskGIT~\citep{chang2022maskgit}, Muse~\citep{chang2023muse}, Show-o~\citep{xie2024show}, and MMaDA~\cite{yang2025mmada} have demonstrated that a discrete diffusion process can generate high-fidelity outputs with great efficiency. Our model's architecture is built upon this discrete diffusion paradigm. We now provide the formal preliminaries, beginning with the foundational forward and reverse processes and culminating in the simplified mask-and-predict training objective that our model employs.

\paragraph{Forward and reverse processes.}
A discrete diffusion model consists of two key processes: (1) The \textit{Forward Process} ($q$), a fixed Markov chain that gradually corrupts input data $\mathbf{x}_0$ over $T$ timesteps into noisy latents $\mathbf{x}_1, \dots, \mathbf{x}_T$; and (2) The \textit{Reverse Process} ($p_\theta$), a learned neural network that reverses this corruption by progressively denoising $\mathbf{x}_T$ to recover the original data distribution. Let's consider a single token $x_0 \in \{1, \dots, K\}$ from a codebook of size $K$. The forward process at each step $t$ is defined by a stochastic transition matrix $\mathbf{Q}_t \in \mathbb{R}^{K \times K}$. A key property is that the distribution of $\mathbf{x}_t$ conditioned on the initial state $\mathbf{x}_0$ is tractable:
\begin{equation}
    q(\mathbf{x}_t | \mathbf{x}_0) = \text{Cat}(\mathbf{x}_t | \mathbf{x}_0 \overline{\mathbf{Q}}_t), \quad \text{where} \quad \overline{\mathbf{Q}}_t = \mathbf{Q}_1 \mathbf{Q}_2 \cdots \mathbf{Q}_t.
\end{equation}
The posterior probability, which is essential for training, is also tractable:
\begin{equation}
    q(\mathbf{x}_{t-1}|\mathbf{x}_{t}, \mathbf{x}_0) = \frac{q(\mathbf{x}_{t}|\mathbf{x}_{t-1})q(\mathbf{x}_{t-1}|\mathbf{x}_0)}{q(\mathbf{x}_{t}| \mathbf{x}_0)} \propto \text{Cat}\left(\mathbf{x}_{t-1} \left| \frac{\mathbf{x}_t\mathbf{Q}_t^{\top} \odot \mathbf{x}_0 \overline{\mathbf{Q}}_{t-1}}{\mathbf{x}_0 \overline{\mathbf{Q}}_{t} \mathbf{x}_t^\top}\right.\right),
\end{equation}
where $\odot$ denotes element-wise product.

\paragraph{Absorbing mask state and transition matrix.}
The design of the transition matrix $\mathbf{Q}_t$ dictates the nature of the corruption. A highly effective approach, inspired by masked language modeling, is to introduce a special \textbf{absorbing \texttt{[MASK]} state}. This expands the token vocabulary to $K+1$ states. Once a token becomes \texttt{[MASK]}, it remains masked for all subsequent timesteps. This explicitly signals corrupted positions to the model. The transition matrix for this "Absorbing-Uniform" process is defined as:
\begin{equation}
    \mathbf{Q}_t=\begin{bmatrix}\omega_t+\nu_t&\nu_t&\cdots&\nu_t&\alpha_t\\\nu_t&\omega_t+\nu_t&\cdots&\nu_t&\alpha_t\\\vdots&\vdots&\ddots&\vdots&\vdots\\\nu_t&\nu_t&\cdots&\omega_t+\nu_t&\alpha_t\\0&0&\cdots&0&1\end{bmatrix} \in \mathbb{R}^{(K+1)\times (K+1)},
\end{equation}
where at each step $t$, a token has a probability $\alpha_t$ to be masked, a probability $\beta_t$ to be replaced by a random token, and a probability $\omega_t = (1-\alpha_t-\beta_t)$ to remain unchanged. The \texttt{[MASK]} token (last row) always transitions to itself.

\paragraph{Objective as mask prediction.}
The training objective for diffusion models is derived by maximizing the Evidence Lower Bound (ELBO) on the data log-likelihood. The negative ELBO, which is minimized during training, can be decomposed into several terms representing different stages of the diffusion process:
\begin{equation}
\begin{aligned}
    -\mathcal{L}_\text{ELBO} = \mathbb{E}_{q} \bigg[ 
        & \underbrace{
            -\log p_{\theta}(\mathbf{x}_0|\mathbf{x}_1)
        }_{\text{Reconstruction Term}} 
        + \sum_{t=2}^T \underbrace{
            \text{KL}(q(\mathbf{x}_{t-1} | \mathbf{x}_t, \mathbf{x}_0) \| p_{\theta}(\mathbf{x}_{t-1}|\mathbf{x}_t))
        }_{\text{Denoising Matching}} \\
        & + \underbrace{
            \text{KL}(q(\mathbf{x}_T | \mathbf{x}_0) \| p(\mathbf{x}_T))
        }_{\text{Prior Matching}} 
    \bigg].
\end{aligned}
\label{eq:elbo_full}
\end{equation}
Here, the objective consists of three main components: (1) a reconstruction term that learns to generate the final data from $\mathbf{x}_1$, (2) a series of KL divergence terms that train the reverse process $p_\theta$ to match the true posterior at each denoising step, and (3) a prior matching term that aligns the final noisy latent with a simple prior distribution. Following derivations in D3PMs~\cite{austin2021structured}, this complex objective can be simplified to a weighted sum of reconstruction terms:
\begin{equation}
    \mathcal{L}_\text{simple} = \sum_{t=1}^T \mathbb{E}_{q(\mathbf{x}_0, \mathbf{x}_t)}[-\log p_\theta(\mathbf{x}_0 | \mathbf{x}_t)].
    \label{eq:elbo_simple}
\end{equation}
When using the absorbing mask state strategy, this simplified objective becomes equivalent to a \textbf{Cross-Entropy loss} for mask token prediction, as used in MaskGIT~\cite{chang2022maskgit}. This approach is highly effective as it focuses the model's capacity on reconstructing only the corrupted parts of the data. Our work leverages this powerful paradigm for both text and image token generation.

\subsection{Group Relative Policy Optimization for Discrete Diffusion Models}
\label{app:diffusion_grpo}

Group Relative Policy Optimization (GRPO)~\citep{guo2025deepseek} is a powerful policy gradient algorithm originally designed for autoregressive models. However, its direct application to discrete diffusion models is non-trivial. The core challenge lies in computing the importance sampling ratios and sequence-level likelihoods; these are straightforward in an autoregressive chain but ill-defined in a non-autoregressive, parallel decoding process. Diffusion models lack a sequential history for token-level probabilities, and their policy distributions are implicitly dependent on masking patterns, making direct likelihood estimation computationally prohibitive.

To bridge this gap, we adopt the efficient random masking framework from MMaDA~\citep{yang2025mmada} to adapt GRPO for our diffusion-based architecture. This strategy circumvents the need for direct likelihood computation by using the model's predictions on randomly masked inputs as an unbiased estimate of the policy likelihoods. First, the advantage $\hat{A}_i$ for each response $o_i$ in a generated group $\{o_j\}_{j=1}^G$ is computed in the standard group-relative manner:
\begin{equation}
    \hat{A}_{i} = \frac{r_i - \text{mean}(\{r_j\}_{j=1}^G)}{\text{std}(\{r_j\}_{j=1}^G) + \epsilon},
\end{equation}
where $r_i$ is the reward for response $o_i$. The policy gradient is then calculated using an importance sampling ratio $r'_{i,t}(\theta)$ defined over a randomly masked version of each response, where a unique mask ratio $p_i \sim U[0,1]$ is sampled for each response at each training step. This allows the standard clipped GRPO objective to be adapted for diffusion models as follows:
\begin{equation}
\begin{aligned}
\mathcal{J}_\text{Diff-GRPO}(\theta)& = \mathbb{E}_{\substack{q\sim\mathcal{D}, \{o_i\}\sim\pi_{\text{old}},\\ \{p_i\}\sim U[0,1]}}
\Bigg[ \frac{1}{G}\sum_{i=1}^{G} \frac{1}{|\text{M}_i|}\sum_{t \in \text{M}_i} \Bigg( 
\min \Big( r'_{i,t}(\theta) \hat{A}_{i}, \\ &
\quad \quad \text{clip} \Big( r'_{i,t}(\theta), 1 - \varepsilon, 1 + \varepsilon \Big) \hat{A}_{i} \Big)
\Bigg) - \beta D_{\text{KL}}(\pi'_{\theta} || \pi'_{\text{ref}})
\Bigg],
\end{aligned}
\end{equation}
where the expectation is also taken over the random mask ratios, the inner summation is only over the masked tokens $\text{M}_i$, and $\pi'$ denotes the policy likelihoods approximated via the masking scheme. This formulation enables stable and efficient policy optimization by effectively adapting the principles of GRPO to a non-autoregressive setting.


\section{Sampling Details on Text and Image}
\label{app:sampling}
\paragraph{Parallel sampling and denoising strategy.}
Our model employs a parallel sampling strategy, predicting logits for all text and image tokens simultaneously in a single forward pass. The denoising process for both modalities is guided by a confidence-based re-masking schedule, inspired by MaskGIT~\citep{chang2022maskgit} and LLaDA~\citep{nie2025large}. Crucially, while the logits are generated jointly, we apply distinct masking schedulers and confidence metrics to the text and image tokens to account for their different statistical properties and generation requirements.

\paragraph{Image token denoising.}
For image generation, we follow the iterative decoding process from MaskGIT. At each timestep $t$, given the current set of $M$ masked image tokens, the model predicts logits $\ell^t = \{\ell_i^t\}_{i=1}^M$. For each masked position $i$, we sample a candidate token $u'_i$ from the predicted probability distribution and compute its confidence score $s_i$. A mask scheduling function $\gamma(t/T)$ determines the number of tokens $m = \lceil\gamma(t/T)M\rceil$ that should be kept (i.e., remain unmasked). We select the $m$ tokens with the highest confidence scores to keep for the next step $t+1$, and the remaining $M-m$ tokens are re-masked. The update rule for a token at position $i$ is:
\begin{equation}
    u_i^{(t+1)} = 
    \begin{cases} 
        u_*, & \text{if } s_i < \text{sorted}_j(s_j)[m] \\
        u'_i, & \text{otherwise}
    \end{cases},
\end{equation}
where $u_*$ represents the \texttt{[MASK]} token and $\text{sorted}_j(s_j)[m]$ is the $m$-th value in the sorted list of confidence scores. This iterative refinement continues until all image tokens are finalized. In our implementation, we generate a 512px image, which is encoded into 1024 discrete tokens and takes 30 steps to decode. 

\paragraph{Text token denoising.}
For text generation, we adopt the semi-autoregressive denoising strategy from LLaDA~\citep{nie2025large}, where the output sequence is generated in blocks from left to right. Within each block, however, generation is non-autoregressive and iterative. The core of this process is a reverse sampling step that transforms a partially masked sequence $\mathbf{x}_t$ at step $t$ into a less masked sequence $\mathbf{x}_s$ at an earlier step $s < t$. This transition is formally characterized by the probability:
\begin{equation}
    q_{s|t}(\mathbf{x}_s|\mathbf{x}_t) = \prod_{i=0}^{N-1} q_{s|t}(x_s^i|\mathbf{x}_t^i) \quad \text{and} \quad q_{s|t}(x_s^i|\mathbf{x}_t^i) = 
    \begin{cases} 
        1, & x_t^i \neq \texttt{[M]}, x_s^i = x_t^i \\
        \frac{1}{1-\alpha_t}, & x_t^i = \texttt{[M]}, x_s^i = \texttt{[M]} \\
        \frac{\alpha_s-\alpha_t}{1-\alpha_t} p_\theta(x_0^i|\mathbf{x}_t), & x_t^i = \texttt{[M]}, x_s^i \neq \texttt{[M]} \\
        0, & \text{otherwise,}
    \end{cases}
\label{eq:text_denoising_transition}
\end{equation}
where $p_\theta(x_0^i|\mathbf{x}_t)$ is the model's prediction of the original token for the masked position $i$ and $\alpha_t = 1 - t$. In practice, this involves an iterative refinement loop. At each step, given the current sequence $\mathbf{x}_t$, we first sample candidate tokens for all masked positions. Then, following the deterministic low-confidence re-masking strategy adopted by LLaDA, we identify the tokens with the lowest prediction confidence scores and re-mask them for the next refinement iteration. 

In our implementation, we generate the sequence with 256 sequence length, in blocks of 64 tokens and 128 steps. At each denoising step within a block, we unmask the two tokens with the lowest confidence scores. This block-based, semi-autoregressive approach is essential for generating coherent and naturally structured sentences, as it mitigates issues like the premature generation of end-of-sequence (\texttt{|EOS|}) tokens that can arise in a fully non-autoregressive setting. 

\section{Details of Training Dataset Curation}
\label{sec:appendix_dataset}
\begin{figure}[ht]
    \centering
    \includegraphics[width=1\linewidth]{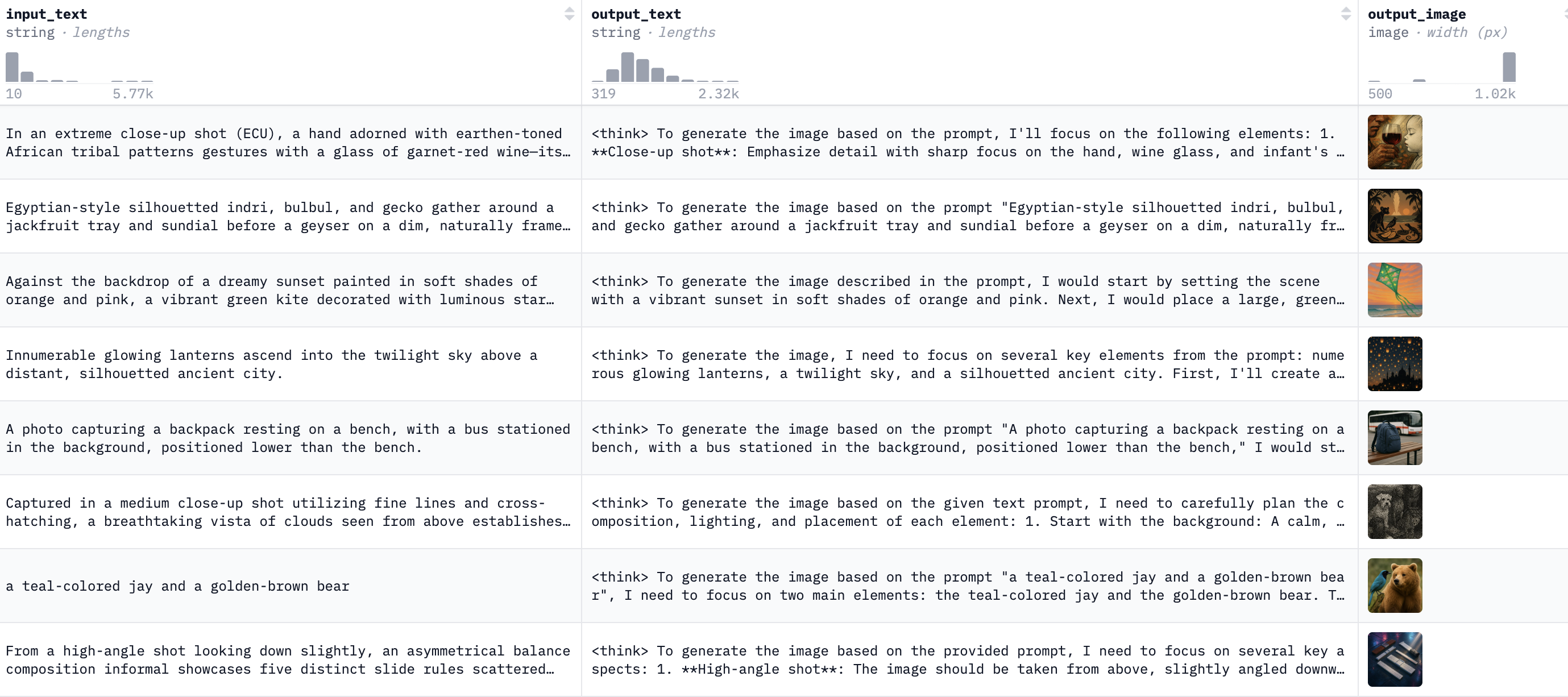}
    \caption{Overview of our dataset for thinking-aware editing}
    \label{fig:app-edit}
\end{figure}
\begin{figure}[ht]
    \centering
    \includegraphics[width=1\linewidth]{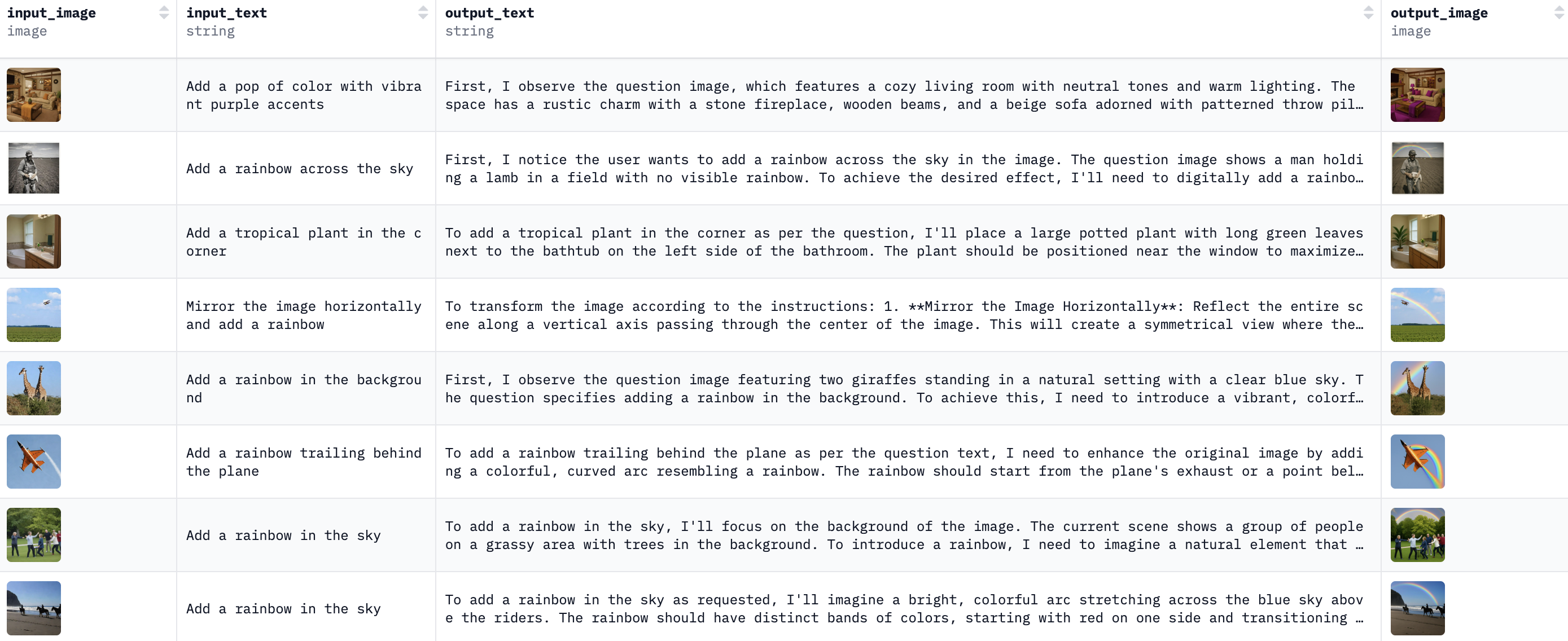}
    \caption{Overview of our dataset for thinking-aware editing}
    \label{fig:app-gen}
\end{figure}
Our training dataset is a carefully curated collection of 150,000 high-quality samples designed for thinking-aware image synthesis. The primary challenge was that existing public datasets for image editing and generation typically provide input-output pairs without the intermediate reasoning traces required by our method. Therefore, our curation process involved three main stages: (1) aggregating data from state-of-the-art sources, (2) generating high-quality reasoning traces to augment this data, and (3) applying a rigorous filtering and enhancement pipeline. The final dataset consists of 100,000 editing pairs and 50,000 generation pairs, achieving a 2:1 ratio. An overview of the dataset is shown in Figure~\ref{fig:app-edit} and ~\ref{fig:app-gen}

\paragraph{Source datasets for editing data.}
We constructed the 100,000 thinking-aware editing pairs by sourcing from four diverse and challenging benchmarks:
\begin{itemize}
    \item \textbf{HQ-Edit}~\citep{hui2024hq}: This dataset provides high-resolution images with a wide variety of detailed editing instructions, serving as a source of high-quality visual content for our training.

    \item \textbf{UltraEdit}~\citep{zhao2024ultraedit}: We leverage UltraEdit for its collection of complex editing instructions that require strong reasoning and compositional abilities, pushing the model beyond simple object manipulation.

    \item \textbf{AnyEdit}~\citep{yu2025anyedit}: Given the vast size of AnyEdit, we selectively sampled from its more challenging categories. Specifically, we focused on the \texttt{implicit\_editing} subset, which contains instructions that do not explicitly mention the target object, requiring the model to infer the user's intent.

    \item \textbf{EditWorld}~\citep{yang2024editworld}: This dataset is crucial for its focus on edits that require world knowledge and complex reasoning, such as causal (e.g., "what if a storm occurs") and temporal (e.g., "What's this man like in twenty years?") edits. To further bolster our model's capabilities in these areas, we performed data augmentation on this subset, using GPT-4o to generate three times the amount of similar, complex reasoning-based instructions and corresponding edits.
\end{itemize}

\paragraph{Source dataset for generation Data.}
For the 50,000 thinking-aware generation pairs, we sourced data from \textbf{ShareGPT4o}~\citep{chen2025sharegpt}. This dataset contains a rich collection of diverse, real-world prompts and corresponding high-quality image outputs, providing a strong foundation for general-purpose, knowledge-intensive image synthesis.

\paragraph{Reasoning trace generation.}
A core step in our curation process was to augment the source data with reasoning traces. Since the original datasets only provide triplets of (`input image`, `instruction`, `output image`), we utilized the powerful multimodal model \textbf{Qwen2.5-VL-7B}~\citep{bai2025qwen2} to generate a plausible reasoning text for each sample. The model was prompted with the input/output image pair and the instruction to produce a step-by-step rationale explaining the transformation. This transformed our dataset into quadruplets: (`input image`, `instruction`, `reasoning trace`, `output image`), which is the required format for our thinking-aware training.

\paragraph{Data filtering and quality control.}
Finally, to ensure the highest quality, we applied a multi-stage filtering pipeline to the entire 150,000-sample dataset. First, we removed near-duplicates to increase data diversity. Second, we used a scoring mechanism based on Qwen-VL to identify and discard samples with low-quality or visually unappealing images. For cases where the instruction was valuable but the image quality was poor, we leveraged \textbf{GPT-4o} to regenerate higher-fidelity candidate images. This comprehensive curation process resulted in a clean, diverse, and high-quality dataset optimized for our training objectives.

\section{Details of ParaBench}
\label{sec:appendix_parabench}

ParaBench is a comprehensive benchmark designed to address the limitations of existing evaluation protocols for thinking-aware image synthesis. Unlike traditional benchmarks that focus solely on the final image, ParaBench is built to assess the entire generation process, including the quality of the intermediate reasoning trace and its synergy with the visual output. It comprises a total of 300 challenging prompts, curated from various sources and divided into 200 for editing and 100 for generation.

\paragraph{Composition of editing prompts.}
The 200 editing prompts are meticulously curated and synthesized from various existing benchmarks to test a wide spectrum of complex reasoning abilities. To provide a structured analysis, we group them into five distinct categories:

\begin{itemize}
    \item \textbf{Spatial Reasoning (40 prompts):} These are tasks requiring a deep understanding of object locations, orientations, and spatial relationships. Examples include instructions like "place the book to the left of the lamp" or "make the person in the background larger."

    \item \textbf{Temporal Reasoning (40 prompts):} These prompts involve reasoning about time and require the model to infer past or future states. Examples include "show what this street might look like 50 years from now" or "revert the shattered vase to its original state."

    \item \textbf{Causal Reasoning (40 prompts):} This category contains instructions that require the model to infer and depict cause-and-effect relationships. Examples include "show the ground after a heavy rain" or "make the plants look like they haven't been watered for weeks."

    \item \textbf{World Knowledge (40 prompts):} These are edits that require external, real-world knowledge to execute correctly. Examples include instructions like "turn this car into a model from the 1980s" or "edit the painting to be in the style of Van Gogh."

    \item \textbf{General Editing (40 prompts):} This category includes a broad set of common, foundational editing operations that do not fit into the specialized categories above. It primarily consists of instructions for adding, removing, or replacing objects and serves as a baseline for fundamental editing capabilities.
\end{itemize}

\paragraph{Composition of generation prompts.}
The 100 generation prompts are sourced from the ShareGPT4o~\citep{chen2025sharegpt} dataset. They are designed to be open-ended and cover a wide range of scenarios, including the generation of creative scenes, complex compositions with multiple interacting objects, and images that require interpreting long, descriptive narratives.

\paragraph{Evaluation axes.}
All 300 prompts in ParaBench are evaluated using our LLM-as-a-judge framework across six fine-grained axes to provide a holistic assessment of a model's performance. The evaluation criteria are as follows:

\begin{itemize}
    \item \textbf{Text Quality:} Assesses the fluency, coherence, and grammatical correctness of the generated reasoning text.
    \item \textbf{Text Alignment:} Measures how well the reasoning text follows the user's input instruction and accurately plans the edit/generation.
    \item \textbf{Image Quality:} Evaluates the photorealism, aesthetic quality, and absence of visual artifacts in the generated image.
    \item \textbf{Image Alignment:} Measures how faithfully the generated image adheres to the user's instruction.
    \item \textbf{Image Consistency (for editing tasks):} Assesses how well the model preserves the unedited parts of the original image, maintaining background, style, and object identity.
    \item \textbf{Output Alignment:} Evaluates the cross-modal consistency between the generated reasoning text and the final generated image.
\end{itemize}
We provide the prompts for thinking-aware image editing in Appendix~\ref{app:prompt}.The prompts for image generation follow the same format, with only minor modifications in the input and representation style.


\section{More Implementation Details}
\label{sec:appendix_implementation}

\paragraph{Training details.}
Our model is initialized from the weights of MMaDA-MixCoT~\citep{yang2025mmada}, which utilizes LLaDA-8B as its text backbone and MagVIT-v2 for image tokenization. The post-training process consists of two stages. In the first stage, we perform supervised finetuning (SFT) for 30,000 steps on our curated dataset of 150,000 thinking-aware samples. In the second stage, we conduct Parallel Reinforcement Learning (ParaRL) for 10,000 steps, using a challenging subset of approximately 15,000 examples (10\%) drawn from the SFT dataset. Both training stages were conducted on 32 NVIDIA A100 GPUs with a global batch size of 768. We utilized the AdamW optimizer with a learning rate of 2e-5 and a cosine learning rate schedule with a warm-up of 500 steps. We drop 10\% of text input and 10\% of image input to support classifier-free guidance sampling.  

In ParaRL, we randomly sample $s=3$ trajectory points. The steps of these certain points are identical in the same rollout and uniformly sampled in all rollouts. We set KL constraints $\beta = 0.0001$ to keep the same with MMaDA's baseline.

\paragraph{Inference details.}
During inference, our model employs a parallel sampling strategy, generating the logits for all text and image tokens simultaneously in a single forward pass. The images are generated with classifier-free guidance scale of 3.5, and text with a scale of 0. 

\section{More Ablation Studies}
\label{app:more_ablation}

\begin{table*}[ht]
\centering
\begin{minipage}{0.48\textwidth}
\centering
\caption{\textbf{Ablation on modality reweighting.}
Default uses $w_{\text{text}}(t){=}1/t$, $w_{\text{img}}(t){=}1$.}
\label{tab:abl_weight}
\resizebox{\textwidth}{!}{
\begin{tabular}{l|ccc}
\toprule
\textbf{Setting} & \textbf{Text Align.} & \textbf{Image Align.} & \textbf{Output Align.} \\
\midrule
Both $1/t$    &   69.5   &   58.1   &   56.3   \\
Both $1$      &   65.7   &   61.9   &   57.0   \\
\rowcolor{mmada_color} $w_{\text{text}}{=}1/t$, $w_{\text{img}}{=}1$ &   \textbf{71}   &   \textbf{63.2}   &  \textbf{59.8}    \\
\bottomrule
\end{tabular}}
\end{minipage}
\hfill
\begin{minipage}{0.48\textwidth}
\centering
\caption{\textbf{Ablation on decoding strategy.}
Fully parallel is our default.}
\label{tab:abl_sampling}
\resizebox{\textwidth}{!}{
\begin{tabular}{l|ccc}
\toprule
\textbf{Strategy} & \textbf{Text Align.} & \textbf{Image Align.} & \textbf{Output Align.} \\
\midrule
Sequential (text $\rightarrow$ image) &  64.2 &  56.5 &  54.1 \\
Semi-parallel (grouped)         &  68.3 &  60.7 &  57.5 \\
\rowcolor{mmada_color} Fully parallel (ours) &  \textbf{71}    &   \textbf{63.2}    &  \textbf{59.8} \\
\bottomrule
\end{tabular}}
\end{minipage}
\end{table*}
We further analyze three key design choices of our framework: 
(1) modality-aware reweighting in the training objective, 
and (2) the decoding strategy (parallel vs semi-parallel vs sequential). 


\paragraph{Modality reweighting.}
Table~\ref{tab:abl_weight} shows that using $w_{\text{text}}(t)=1/t$ and $w_{\text{img}}(t)=1$ stabilizes image training and yields the best overall performance. 
Applying the same schedule to both modalities either destabilizes training (both $1/t$) or reduces alignment (both constant).

\paragraph{Decoding strategy.}
Table~\ref{tab:abl_sampling} contrasts fully parallel, semi-parallel, and fully sequential decoding. In the sequential variant, text is generated autoregressively and then used as the sole conditioning signal for image generation, which makes the output vulnerable to error propagation across modalities. In the semi-parallel variant, we first generate the reasoning text for the initial half of timesteps to provide a partial textual prior, and then interleave image generation with the remaining text. This strategy mitigates some sequential errors and yields improvements over the fully sequential baseline. Finally, the fully parallel variant, i.e., \method, generates text and image jointly at every denoising step. We find that fully parallel decoding achieves strong results without requiring extensive textual priors, likely because the early image steps can already establish coarse scene layouts, and excessive initial text may even bias attention toward irrelevant details.

\section{Limitations and Future Work}

Although our approach achieves notable improvements, several limitations remain. First, our base model MMaDA is trained on relatively limited data, which constrains its fundamental capabilities. As a result, it is difficult to consistently surpass large-scale models such as Bagel that benefit from substantially larger training corpora. Second, our current sampling and training strategies are not yet fully unified across modalities, and exploring more integrated interaction paradigms may further enhance performance.  

For future work, we plan to extend our paradigm to broader scenarios, such as story generation and multimodal outputs that combine text and images, which we believe will further demonstrate the potential of parallel thinking-aware generation.  
\section{Prompts for evaluation}
\label{app:prompt}

\clearpage
\begin{figure}[ht]
    \centering
    \includegraphics[width=1\linewidth]{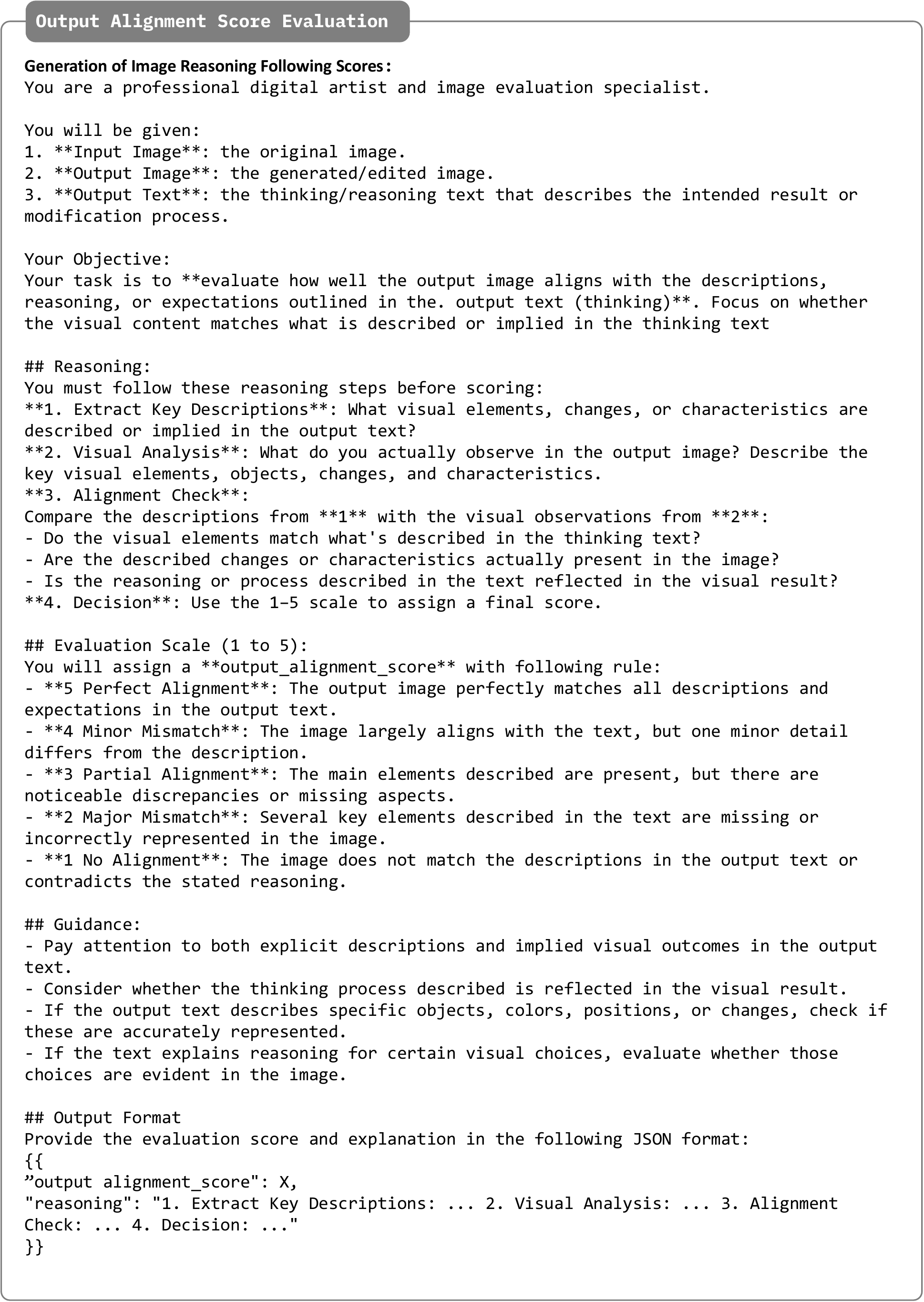}
    \caption{Output alignment evaluation prompt}
    \label{fig:OA}
\end{figure}
\clearpage
\begin{figure}[ht]
    \centering
    \includegraphics[width=1\linewidth]{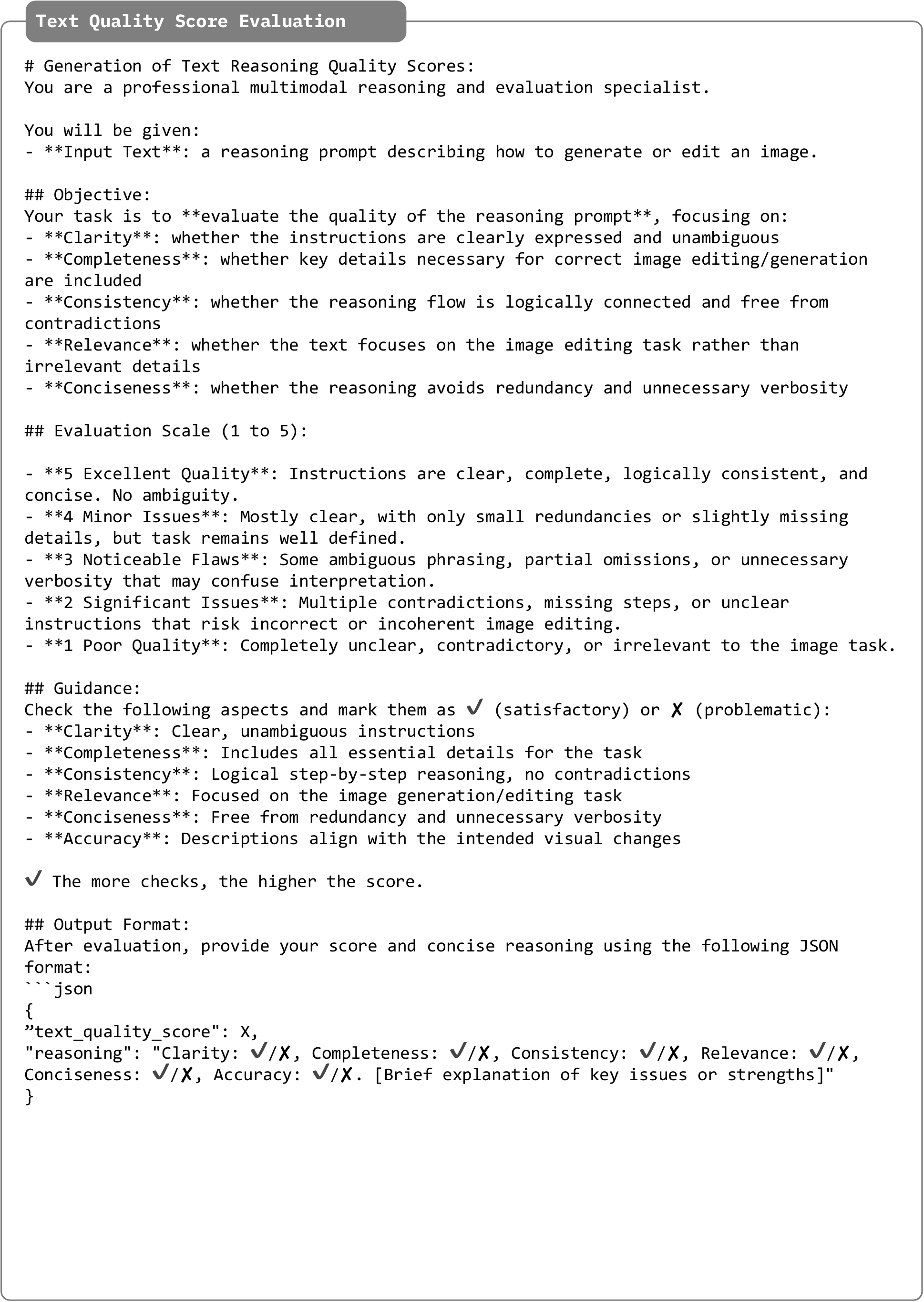}
    \caption{Text quality evaluation prompt}
    \label{fig:p1}
\end{figure}
\clearpage
\begin{figure}[ht]
    \centering
    \includegraphics[width=1\linewidth]{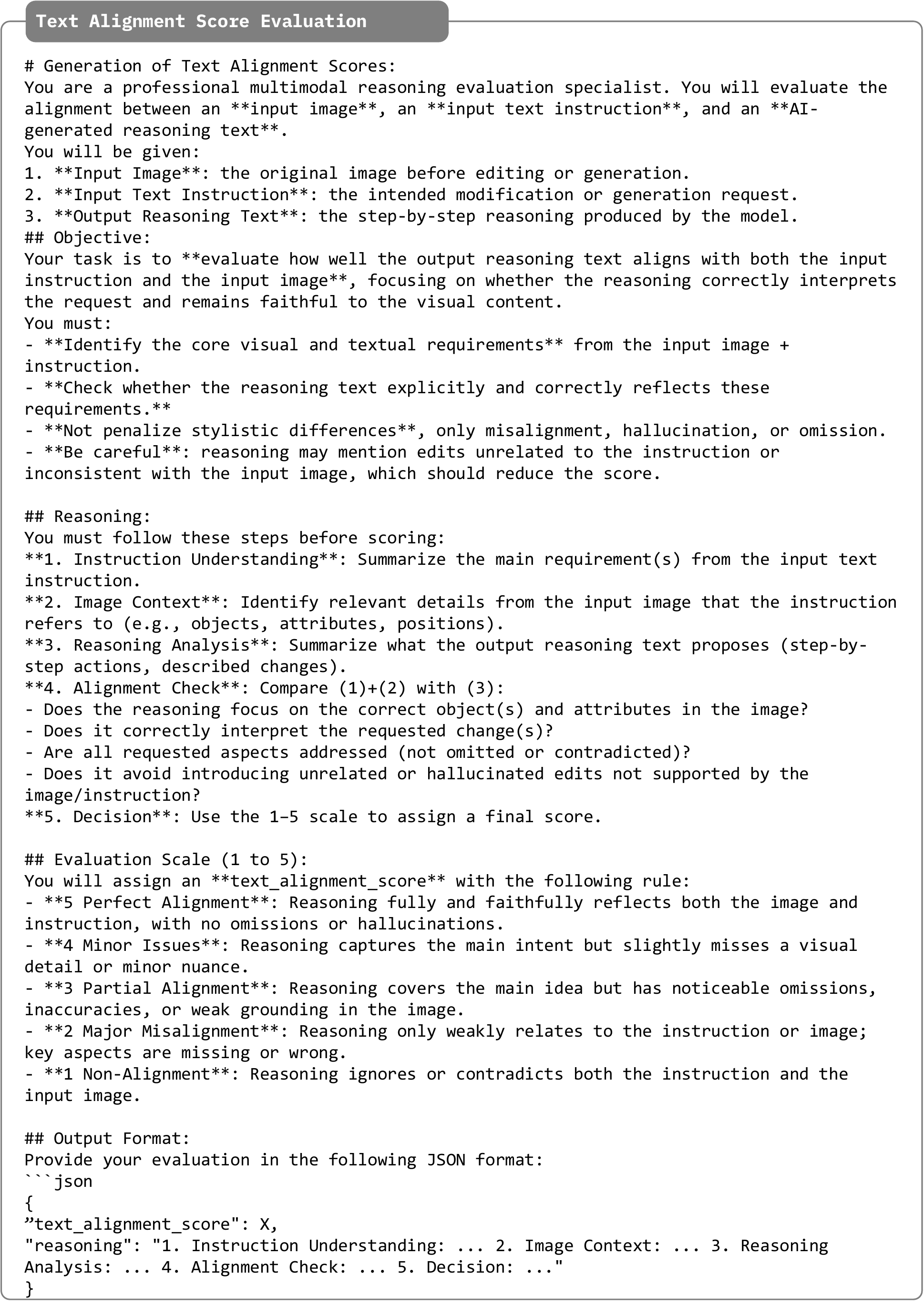}
    \caption{Text alignment evaluation prompt}
    \label{fig:p2}
\end{figure}
\clearpage
\begin{figure}[ht]
    \centering
    \includegraphics[width=1\linewidth]{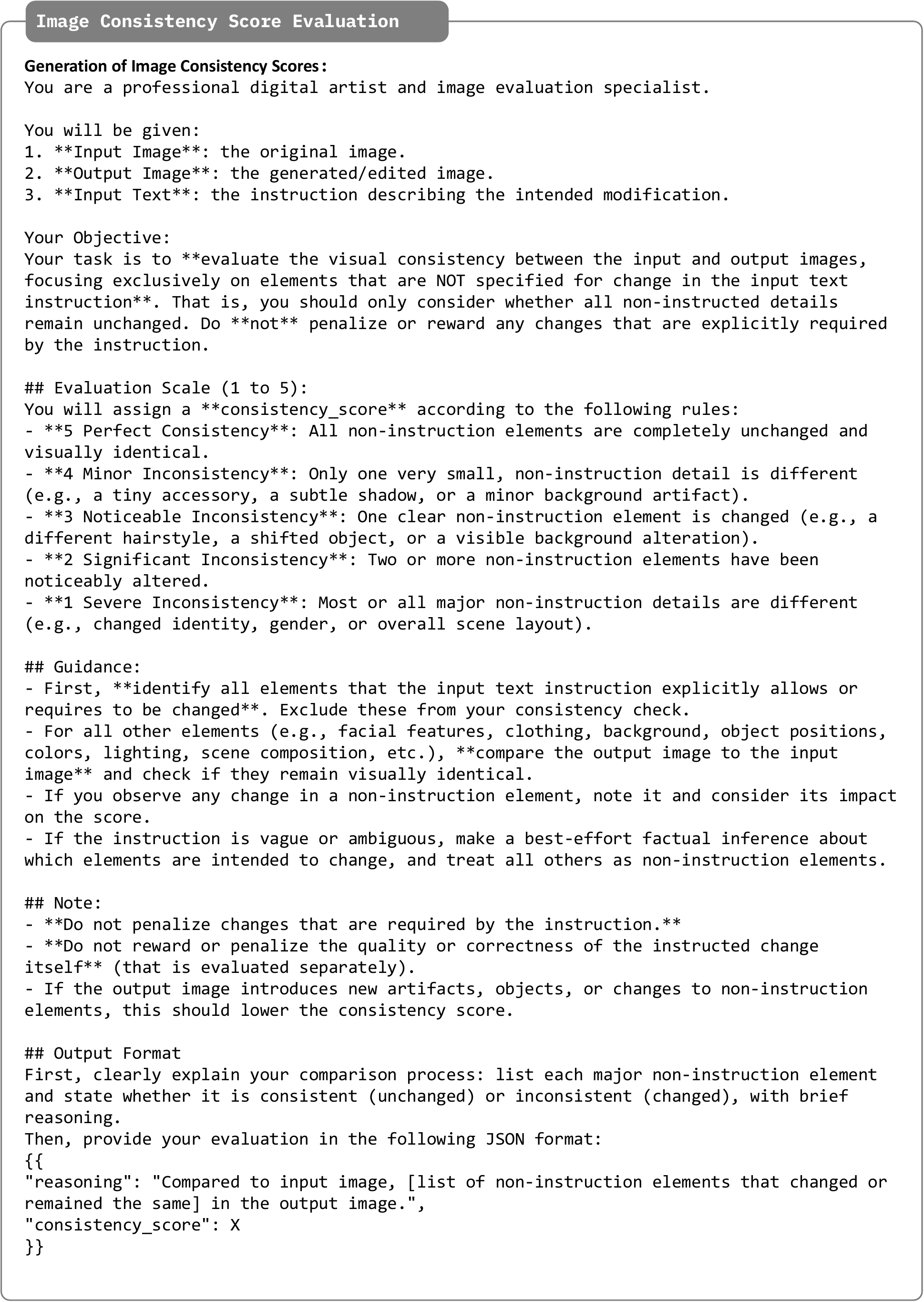}
    \caption{Image consistency evaluation prompt}
    \label{fig:p3}
\end{figure}
\clearpage
\begin{figure}[ht]
    \centering
    \includegraphics[width=1\linewidth]{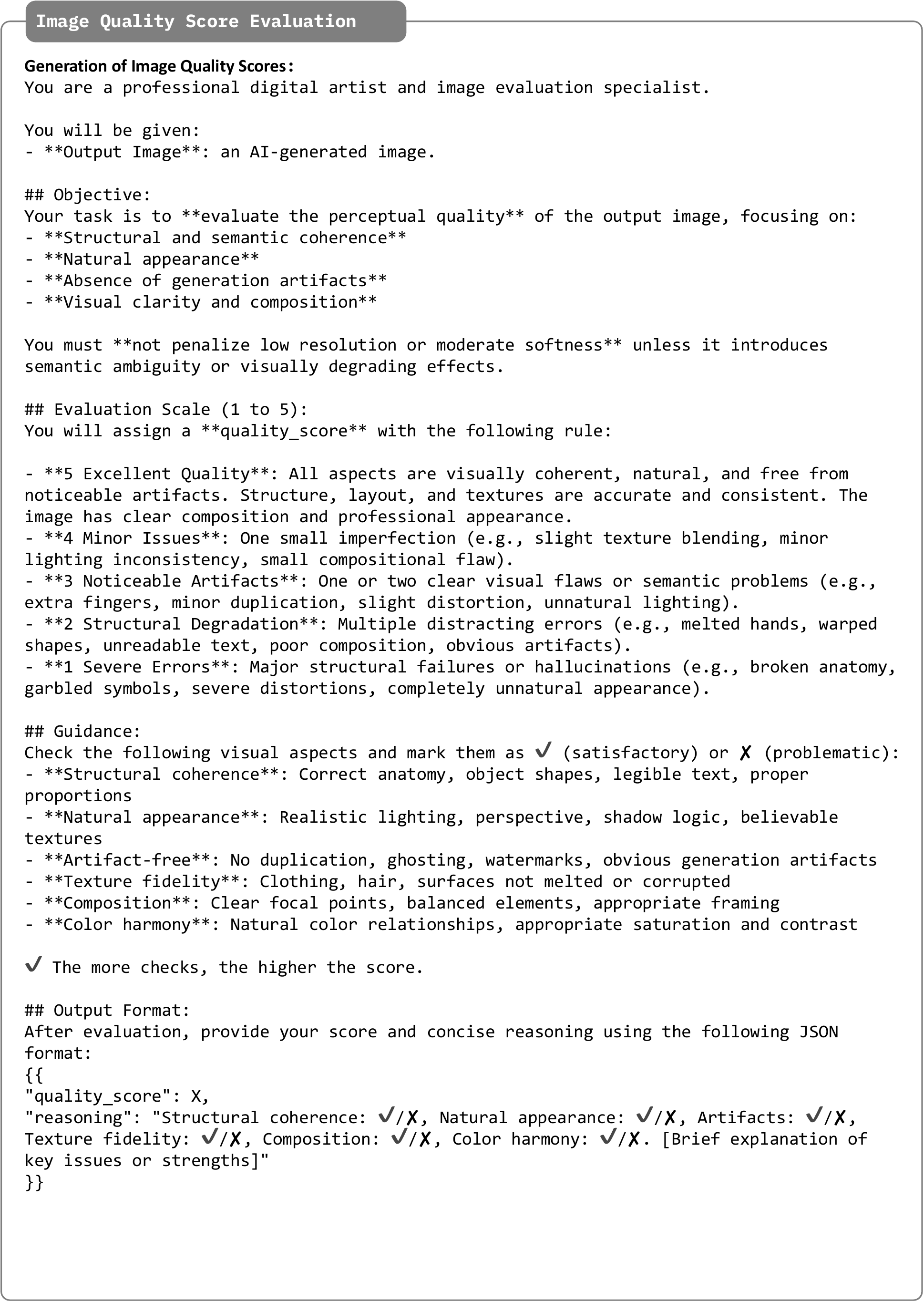}
    \caption{Image quality evaluation prompt}
    \label{fig:p4}
\end{figure}
\clearpage
\begin{figure}[ht]
    \centering
    \includegraphics[width=1\linewidth]{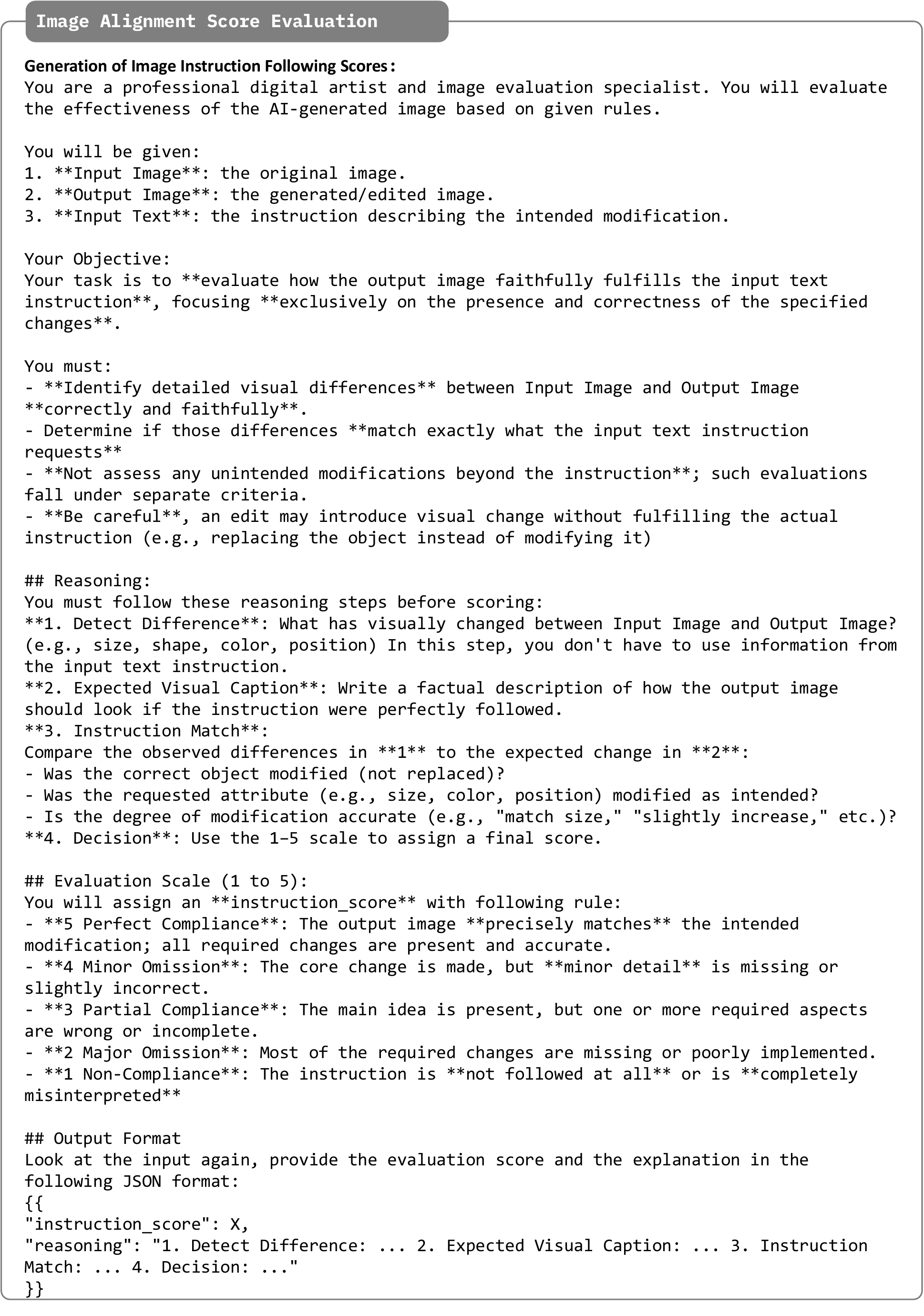}
    \caption{Image alignment evaluation prompt}
    \label{fig:p5}
\end{figure}

\clearpage
\newpage

\bibliography{iclr2026_conference}

@inproceedings{esser2024scaling,
  title={Scaling rectified flow transformers for high-resolution image synthesis},
  author={Esser, Patrick and Kulal, Sumith and Blattmann, Andreas and Entezari, Rahim and M{\"u}ller, Jonas and Saini, Harry and Levi, Yam and Lorenz, Dominik and Sauer, Axel and Boesel, Frederic and others},
  booktitle={Forty-first international conference on machine learning},
  year={2024}
}

@misc{flux,
  title = {FLUX},
  author = {Black Forest Labs},
  year = {2024},
  url = {https://github.com/black-forest-labs/flux}
}

@inproceedings{wei2024omniedit,
  title={Omniedit: Building image editing generalist models through specialist supervision},
  author={Wei, Cong and Xiong, Zheyang and Ren, Weiming and Du, Xeron and Zhang, Ge and Chen, Wenhu},
  booktitle={ICLR},
  year={2024}
}

@article{liu2025step1xeditpracticalframeworkgeneral,
      title={Step1X-Edit: A Practical Framework for General Image Editing}, 
      author={Shiyu Liu and Yucheng Han and Peng Xing and Fukun Yin and Rui Wang and Wei Cheng and Jiaqi Liao and Yingming Wang and Honghao Fu and Chunrui Han and Guopeng Li and Yuang Peng and Quan Sun and Jingwei Wu and Yan Cai and Zheng Ge and Ranchen Ming and Lei Xia and Xianfang Zeng and Yibo Zhu and Binxing Jiao and Xiangyu Zhang and Gang Yu and Daxin Jiang},
      year={2025},
    journal={arXiv preprint arXiv:2504.17761}
}

@article{jiang2025t2i,
  title={T2i-r1: Reinforcing image generation with collaborative semantic-level and token-level cot},
  author={Jiang, Dongzhi and Guo, Ziyu and Zhang, Renrui and Zong, Zhuofan and Li, Hao and Zhuo, Le and Yan, Shilin and Heng, Pheng-Ann and Li, Hongsheng},
  journal={arXiv preprint arXiv:2505.00703},
  year={2025}
}

@article{deng2025emerging,
  title={Emerging properties in unified multimodal pretraining},
  author={Deng, Chaorui and Zhu, Deyao and Li, Kunchang and Gou, Chenhui and Li, Feng and Wang, Zeyu and Zhong, Shu and Yu, Weihao and Nie, Xiaonan and Song, Ziang and others},
  journal={arXiv preprint arXiv:2505.14683},
  year={2025}
}

@article{wu2025kris,
  title={KRIS-Bench: Benchmarking Next-Level Intelligent Image Editing Models},
  author={Wu, Yongliang and Li, Zonghui and Hu, Xinting and Ye, Xinyu and Zeng, Xianfang and Yu, Gang and Zhu, Wenbo and Schiele, Bernt and Yang, Ming-Hsuan and Yang, Xu},
  journal={arXiv preprint arXiv:2505.16707},
  year={2025}
}

@article{zhao2025envisioning,
  title={Envisioning beyond the pixels: Benchmarking reasoning-informed visual editing},
  author={Zhao, Xiangyu and Zhang, Peiyuan and Tang, Kexian and Zhu, Xiaorong and Li, Hao and Chai, Wenhao and Zhang, Zicheng and Xia, Renqiu and Zhai, Guangtao and Yan, Junchi and others},
  journal={arXiv preprint arXiv:2504.02826},
  year={2025}
}

@article{niu2025wise,
  title={Wise: A world knowledge-informed semantic evaluation for text-to-image generation},
  author={Niu, Yuwei and Ning, Munan and Zheng, Mengren and Jin, Weiyang and Lin, Bin and Jin, Peng and Liao, Jiaqi and Feng, Chaoran and Ning, Kunpeng and Zhu, Bin and others},
  journal={arXiv preprint arXiv:2503.07265},
  year={2025}
}

@article{nie2025large,
  title={Large language diffusion models},
  author={Nie, Shen and Zhu, Fengqi and You, Zebin and Zhang, Xiaolu and Ou, Jingyang and Hu, Jun and Zhou, Jun and Lin, Yankai and Wen, Ji-Rong and Li, Chongxuan},
  journal={arXiv preprint arXiv:2502.09992},
  year={2025}
}

@article{yang2025mmada,
  title={Mmada: Multimodal large diffusion language models},
  author={Yang, Ling and Tian, Ye and Li, Bowen and Zhang, Xinchen and Shen, Ke and Tong, Yunhai and Wang, Mengdi},
  journal={arXiv preprint arXiv:2505.15809},
  year={2025}
}

@misc{dream2025,
    title = {Dream 7B},
    url = {https://hkunlp.github.io/blog/2025/dream},
    author = {Ye, Jiacheng and Xie, Zhihui and Zheng, Lin and Gao, Jiahui and Wu, Zirui and Jiang, Xin and Li, Zhenguo and Kong, Lingpeng},
    year = {2025}
}

@article{team2024chameleon,
  title={Chameleon: Mixed-modal early-fusion foundation models},
  author={Team, Chameleon},
  journal={arXiv preprint arXiv:2405.09818},
  year={2024}
}

@article{liao2025mogao,
  title={Mogao: An omni foundation model for interleaved multi-modal generation},
  author={Liao, Chao and Liu, Liyang and Wang, Xun and Luo, Zhengxiong and Zhang, Xinyu and Zhao, Wenliang and Wu, Jie and Li, Liang and Tian, Zhi and Huang, Weilin},
  journal={arXiv preprint arXiv:2505.05472},
  year={2025}
}

@article{wu2025omnigen2,
  title={OmniGen2: Exploration to Advanced Multimodal Generation},
  author={Wu, Chenyuan and Zheng, Pengfei and Yan, Ruiran and Xiao, Shitao and Luo, Xin and Wang, Yueze and Li, Wanli and Jiang, Xiyan and Liu, Yexin and Zhou, Junjie and others},
  journal={arXiv preprint arXiv:2506.18871},
  year={2025}
}

@article{huang2025interleaving,
  title={Interleaving Reasoning for Better Text-to-Image Generation},
  author={Huang, Wenxuan and Chen, Shuang and Xie, Zheyong and Cao, Shaosheng and Tang, Shixiang and Shen, Yufan and Yin, Qingyu and Hu, Wenbo and Wang, Xiaoman and Tang, Yuntian and others},
  journal={arXiv preprint arXiv:2509.06945},
  year={2025}
}

@article{fang2025got,
  title={Got: Unleashing reasoning capability of multimodal large language model for visual generation and editing},
  author={Fang, Rongyao and Duan, Chengqi and Wang, Kun and Huang, Linjiang and Li, Hao and Yan, Shilin and Tian, Hao and Zeng, Xingyu and Zhao, Rui and Dai, Jifeng and others},
  journal={arXiv preprint arXiv:2503.10639},
  year={2025}
}

@article{guo2025can,
  title={Can We Generate Images with CoT? Let's Verify and Reinforce Image Generation Step by Step},
  author={Guo, Ziyu and Zhang, Renrui and Tong, Chengzhuo and Zhao, Zhizheng and Huang, Rui and Zhang, Haoquan and Zhang, Manyuan and Liu, Jiaming and Zhang, Shanghang and Gao, Peng and others},
  journal={arXiv preprint arXiv:2501.13926},
  year={2025}
}

@article{llada1.5,
  title={LLaDA 1.5: Variance-Reduced Preference Optimization for Large Language Diffusion Models},
  author={Zhu, Fengqi and Wang, Rongzhen and Nie, Shen and Zhang, Xiaolu and Wu, Chunwei and Hu, Jun and Zhou, Jun and Chen, Jianfei and Lin, Yankai and Wen, Ji-Rong and others},
  journal={arXiv preprint arXiv:2505.19223},
  year={2025}
}

@article{lladav,
  title={Llada-v: Large language diffusion models with visual instruction tuning},
  author={You, Zebin and Nie, Shen and Zhang, Xiaolu and Hu, Jun and Zhou, Jun and Lu, Zhiwu and Wen, Ji-Rong and Li, Chongxuan},
  journal={arXiv preprint arXiv:2505.16933},
  year={2025}
}

@article{ddpm,
  title={Denoising diffusion probabilistic models},
  author={Ho, Jonathan and Jain, Ajay and Abbeel, Pieter},
  journal={Advances in neural information processing systems},
  volume={33},
  pages={6840--6851},
  year={2020}
}

@article{ddim,
  title={Denoising diffusion implicit models},
  author={Song, Jiaming and Meng, Chenlin and Ermon, Stefano},
  journal={arXiv preprint arXiv:2010.02502},
  year={2020}
}

@inproceedings{sd,
  title={High-resolution image synthesis with latent diffusion models},
  author={Rombach, Robin and Blattmann, Andreas and Lorenz, Dominik and Esser, Patrick and Ommer, Bj{\"o}rn},
  booktitle={Proceedings of the IEEE/CVF conference on computer vision and pattern recognition},
  pages={10684--10695},
  year={2022}
}

@inproceedings{dit,
  title={Scalable diffusion models with transformers},
  author={Peebles, William and Xie, Saining},
  booktitle={Proceedings of the IEEE/CVF international conference on computer vision},
  pages={4195--4205},
  year={2023}
}

@inproceedings{sd3,
  title={Scaling rectified flow transformers for high-resolution image synthesis},
  author={Esser, Patrick and Kulal, Sumith and Blattmann, Andreas and Entezari, Rahim and M{\"u}ller, Jonas and Saini, Harry and Levi, Yam and Lorenz, Dominik and Sauer, Axel and Boesel, Frederic and others},
  booktitle={Forty-first international conference on machine learning},
  year={2024}
}

@article{analog,
  title={Analog bits: Generating discrete data using diffusion models with self-conditioning},
  author={Chen, Ting and Zhang, Ruixiang and Hinton, Geoffrey},
  journal={arXiv preprint arXiv:2208.04202},
  year={2022}
}

@article{tess,
  title={Tess: Text-to-text self-conditioned simplex diffusion},
  author={Mahabadi, Rabeeh Karimi and Ivison, Hamish and Tae, Jaesung and Henderson, James and Beltagy, Iz and Peters, Matthew E and Cohan, Arman},
  journal={arXiv preprint arXiv:2305.08379},
  year={2023}
}

@article{dinoiser,
  title={Dinoiser: Diffused conditional sequence learning by manipulating noises},
  author={Ye, Jiasheng and Zheng, Zaixiang and Bao, Yu and Qian, Lihua and Wang, Mingxuan},
  journal={arXiv preprint arXiv:2302.10025},
  year={2023}
}

@article{diffuseq,
  title={Diffuseq: Sequence to sequence text generation with diffusion models},
  author={Gong, Shansan and Li, Mukai and Feng, Jiangtao and Wu, Zhiyong and Kong, LingPeng},
  journal={arXiv preprint arXiv:2210.08933},
  year={2022}
}

@article{scalingdiscrete,
  title={Scaling diffusion language models via adaptation from autoregressive models},
  author={Gong, Shansan and Agarwal, Shivam and Zhang, Yizhe and Ye, Jiacheng and Zheng, Lin and Li, Mukai and An, Chenxin and Zhao, Peilin and Bi, Wei and Han, Jiawei and others},
  journal={arXiv preprint arXiv:2410.17891},
  year={2024}
}

@article{yourdiscrete,
  title={Your absorbing discrete diffusion secretly models the conditional distributions of clean data},
  author={Ou, Jingyang and Nie, Shen and Xue, Kaiwen and Zhu, Fengqi and Sun, Jiacheng and Li, Zhenguo and Li, Chongxuan},
  journal={arXiv preprint arXiv:2406.03736},
  year={2024}
}

@article{longllada,
  title={LongLLaDA: Unlocking Long Context Capabilities in Diffusion LLMs},
  author={Liu, Xiaoran and Liu, Zhigeng and Huang, Zengfeng and Guo, Qipeng and He, Ziwei and Qiu, Xipeng},
  journal={arXiv preprint arXiv:2506.14429},
  year={2025}
}

@article{dream7b,
  title={Dream 7B: Diffusion Large Language Models},
  author={Ye, Jiacheng and Xie, Zhihui and Zheng, Lin and Gao, Jiahui and Wu, Zirui and Jiang, Xin and Li, Zhenguo and Kong, Lingpeng},
  journal={arXiv preprint arXiv:2508.15487},
  year={2025}
}

@article{lavida,
  title={Lavida: A large diffusion language model for multimodal understanding},
  author={Li, Shufan and Kallidromitis, Konstantinos and Bansal, Hritik and Gokul, Akash and Kato, Yusuke and Kozuka, Kazuki and Kuen, Jason and Lin, Zhe and Chang, Kai-Wei and Grover, Aditya},
  journal={arXiv preprint arXiv:2505.16839},
  year={2025}
}

@misc{chen2025r1v,
  author       = {Chen, Liang and Li, Lei and Zhao, Haozhe and Song, Yifan and Vinci},
  title        = {R1-V: Reinforcing Super Generalization Ability in Vision-Language Models with Less Than \$3},
  howpublished = {\url{https://github.com/Deep-Agent/R1-V}},
  note         = {Accessed: 2025-02-02},
  year         = {2025}
}

@article{meng2025mm,
  title={MM-Eureka: Exploring Visual Aha Moment with Rule-based Large-scale Reinforcement Learning},
  author={Meng, Fanqing and Du, Lingxiao and Liu, Zongkai and Zhou, Zhixiang and Lu, Quanfeng and Fu, Daocheng and Shi, Botian and Wang, Wenhai and He, Junjun and Zhang, Kaipeng and others},
  journal={arXiv preprint arXiv:2503.07365},
  year={2025}
}

@article{r1vl,
  title={R1-VL: Learning to Reason with Multimodal Large Language Models via Step-wise Group Relative Policy Optimization},
  author={Zhang, Jingyi and Huang, Jiaxing and Yao, Huanjin and Liu, Shunyu and Zhang, Xikun and Lu, Shijian and Tao, Dacheng},
  journal={arXiv preprint arXiv:2503.12937},
  year={2025}
}

@article{r1-onevision,
  title={R1-Onevision: Advancing Generalized Multimodal Reasoning through Cross-Modal Formalization},
  author={Yang, Yi and He, Xiaoxuan and Pan, Hongkun and Jiang, Xiyan and Deng, Yan and Yang, Xingtao and Lu, Haoyu and Yin, Dacheng and Rao, Fengyun and Zhu, Minfeng and others},
  journal={arXiv preprint arXiv:2503.10615},
  year={2025}
}

@article{visionr1,
  title={Vision-r1: Incentivizing reasoning capability in multimodal large language models},
  author={Huang, Wenxuan and Jia, Bohan and Zhai, Zijie and Cao, Shaosheng and Ye, Zheyu and Zhao, Fei and Hu, Yao and Lin, Shaohui},
  journal={arXiv preprint arXiv:2503.06749},
  year={2025}
}

@misc{openvlthinker,
      title={OpenVLThinker: An Early Exploration to Complex Vision-Language Reasoning via Iterative Self-Improvement}, 
      author={Yihe Deng and Hritik Bansal and Fan Yin and Nanyun Peng and Wei Wang and Kai-Wei Chang},
      year={2025},
      eprint={2503.17352},
      archivePrefix={arXiv},
      primaryClass={cs.CV},
      url={https://arxiv.org/abs/2503.17352}, 
}

@article{liu2025seg,
  title={Seg-zero: Reasoning-chain guided segmentation via cognitive reinforcement},
  author={Liu, Yuqi and Peng, Bohao and Zhong, Zhisheng and Yue, Zihao and Lu, Fanbin and Yu, Bei and Jia, Jiaya},
  journal={arXiv preprint arXiv:2503.06520},
  year={2025}
}

@article{guo2025deepseek,
  title={DeepSeek-R1: Incentivizing Reasoning Capability in LLMs via Reinforcement Learning},
  author={Guo, Daya and Yang, Dejian and Zhang, Haowei and Song, Junxiao and Zhang, Ruoyu and Xu, Runxin and Zhu, Qihao and Ma, Shirong and Wang, Peiyi and Bi, Xiao and others},
  journal={arXiv preprint arXiv:2501.12948},
  year={2025}
}

@article{jiang2025co,
  title={Co-Reinforcement Learning for Unified Multimodal Understanding and Generation},
  author={Jiang, Jingjing and Si, Chongjie and Luo, Jun and Zhang, Hanwang and Ma, Chao},
  journal={arXiv preprint arXiv:2505.17534},
  year={2025}
}

@article{hong2025reinforcing,
  title={Reinforcing Multimodal Understanding and Generation with Dual Self-rewards},
  author={Hong, Jixiang and Zhang, Yiran and Wang, Guanzhong and Liu, Yi and Wen, Ji-Rong and Yan, Rui},
  journal={arXiv preprint arXiv:2506.07963},
  year={2025}
}

@article{gong2025diffucoder,
  title={DiffuCoder: Understanding and Improving Masked Diffusion Models for Code Generation},
  author={Gong, Shansan and Zhang, Ruixiang and Zheng, Huangjie and Gu, Jiatao and Jaitly, Navdeep and Kong, Lingpeng and Zhang, Yizhe},
  journal={arXiv preprint arXiv:2506.20639},
  year={2025}
}

@article{bai2024meissonic,
  title={Meissonic: Revitalizing masked generative transformers for efficient high-resolution text-to-image synthesis},
  author={Bai, Jinbin and Ye, Tian and Chow, Wei and Song, Enxin and Li, Xiangtai and Dong, Zhen and Zhu, Lei and Yan, Shuicheng},
  journal={arXiv preprint arXiv:2410.08261},
  year={2024}
}

@article{xie2024show,
  title={Show-o: One single transformer to unify multimodal understanding and generation},
  author={Xie, Jinheng and Mao, Weijia and Bai, Zechen and Zhang, David Junhao and Wang, Weihao and Lin, Kevin Qinghong and Gu, Yuchao and Chen, Zhijie and Yang, Zhenheng and Shou, Mike Zheng},
  journal={arXiv preprint arXiv:2408.12528},
  year={2024}
}

@article{magvitv2,
  title={Language Model Beats Diffusion--Tokenizer is Key to Visual Generation},
  author={Yu, Lijun and Lezama, Jos{\'e} and Gundavarapu, Nitesh B and Versari, Luca and Sohn, Kihyuk and Minnen, David and Cheng, Yong and Gupta, Agrim and Gu, Xiuye and Hauptmann, Alexander G and others},
  journal={arXiv preprint arXiv:2310.05737},
  year={2023}
}

@article{li2024process,
  title={Process reward model with q-value rankings},
  author={Li, Wendi and Li, Yixuan},
  journal={arXiv preprint arXiv:2410.11287},
  year={2024}
}

@article{wang2025revolutionizing,
  title={Revolutionizing reinforcement learning framework for diffusion large language models},
  author={Wang, Yinjie and Yang, Ling and Li, Bowen and Tian, Ye and Shen, Ke and Wang, Mengdi},
  journal={arXiv preprint arXiv:2509.06949},
  year={2025}
}

@article{zhao2024ultraedit,
  title={Ultraedit: Instruction-based fine-grained image editing at scale},
  author={Zhao, Haozhe and Ma, Xiaojian Shawn and Chen, Liang and Si, Shuzheng and Wu, Rujie and An, Kaikai and Yu, Peiyu and Zhang, Minjia and Li, Qing and Chang, Baobao},
  journal={Advances in Neural Information Processing Systems},
  volume={37},
  pages={3058--3093},
  year={2024}
}

@inproceedings{yu2025anyedit,
  title={Anyedit: Mastering unified high-quality image editing for any idea},
  author={Yu, Qifan and Chow, Wei and Yue, Zhongqi and Pan, Kaihang and Wu, Yang and Wan, Xiaoyang and Li, Juncheng and Tang, Siliang and Zhang, Hanwang and Zhuang, Yueting},
  booktitle={Proceedings of the Computer Vision and Pattern Recognition Conference},
  pages={26125--26135},
  year={2025}
}

@article{hui2024hq,
  title={Hq-edit: A high-quality dataset for instruction-based image editing},
  author={Hui, Mude and Yang, Siwei and Zhao, Bingchen and Shi, Yichun and Wang, Heng and Wang, Peng and Zhou, Yuyin and Xie, Cihang},
  journal={arXiv preprint arXiv:2404.09990},
  year={2024}
}

@article{bai2025qwen2,
  title={Qwen2. 5-vl technical report},
  author={Bai, Shuai and Chen, Keqin and Liu, Xuejing and Wang, Jialin and Ge, Wenbin and Song, Sibo and Dang, Kai and Wang, Peng and Wang, Shijie and Tang, Jun and others},
  journal={arXiv preprint arXiv:2502.13923},
  year={2025}
}

@article{yang2024editworld,
  title={Editworld: Simulating world dynamics for instruction-following image editing},
  author={Yang, Ling and Zeng, Bohan and Liu, Jiaming and Li, Hong and Xu, Minghao and Zhang, Wentao and Yan, Shuicheng},
  journal={arXiv preprint arXiv:2405.14785},
  year={2024}
}

@article{chen2025sharegpt,
  title={ShareGPT-4o-Image: Aligning Multimodal Models with GPT-4o-Level Image Generation},
  author={Chen, Junying and Cai, Zhenyang and Chen, Pengcheng and Chen, Shunian and Ji, Ke and Wang, Xidong and Yang, Yunjin and Wang, Benyou},
  journal={arXiv preprint arXiv:2506.18095},
  year={2025}
}

@article{wu2025qwen,
  title={Qwen-image technical report},
  author={Wu, Chenfei and Li, Jiahao and Zhou, Jingren and Lin, Junyang and Gao, Kaiyuan and Yan, Kun and Yin, Sheng-ming and Bai, Shuai and Xu, Xiao and Chen, Yilei and others},
  journal={arXiv preprint arXiv:2508.02324},
  year={2025}
}

@misc{labs2025flux1kontextflowmatching,
      title={FLUX.1 Kontext: Flow Matching for In-Context Image Generation and Editing in Latent Space},
      author={Black Forest Labs and Stephen Batifol and Andreas Blattmann and Frederic Boesel and Saksham Consul and Cyril Diagne and Tim Dockhorn and Jack English and Zion English and Patrick Esser and Sumith Kulal and Kyle Lacey and Yam Levi and Cheng Li and Dominik Lorenz and Jonas Müller and Dustin Podell and Robin Rombach and Harry Saini and Axel Sauer and Luke Smith},
      year={2025},
      eprint={2506.15742},
      archivePrefix={arXiv},
      primaryClass={cs.GR},
      url={https://arxiv.org/abs/2506.15742},
}

@article{ghosh2023geneval,
  title={Geneval: An object-focused framework for evaluating text-to-image alignment},
  author={Ghosh, Dhruba and Hajishirzi, Hannaneh and Schmidt, Ludwig},
  journal={Advances in Neural Information Processing Systems},
  volume={36},
  pages={52132--52152},
  year={2023}
}

@article{liu2025flow,
  title={Flow-grpo: Training flow matching models via online rl},
  author={Liu, Jie and Liu, Gongye and Liang, Jiajun and Li, Yangguang and Liu, Jiaheng and Wang, Xintao and Wan, Pengfei and Zhang, Di and Ouyang, Wanli},
  journal={arXiv preprint arXiv:2505.05470},
  year={2025}
}

@inproceedings{vqdiffusion,
  title={Vector quantized diffusion model for text-to-image synthesis},
  author={Gu, Shuyang and Chen, Dong and Bao, Jianmin and Wen, Fang and Zhang, Bo and Chen, Dongdong and Yuan, Lu and Guo, Baining},
  booktitle={CVPR},
  pages={10696--10706},
  year={2022}
}

@inproceedings{chang2022maskgit,
  title={Maskgit: Masked generative image transformer},
  author={Chang, Huiwen and Zhang, Han and Jiang, Lu and Liu, Ce and Freeman, William T},
  booktitle={CVPR},
  pages={11315--11325},
  year={2022}
}

@article{chang2023muse,
  title={Muse: Text-to-image generation via masked generative transformers},
  author={Chang, Huiwen and Zhang, Han and Barber, Jarred and Maschinot, AJ and Lezama, Jose and Jiang, Lu and Yang, Ming-Hsuan and Murphy, Kevin and Freeman, William T and Rubinstein, Michael and others},
  journal={arXiv preprint arXiv:2301.00704},
  year={2023}
}

@article{austin2021structured,
  title={Structured denoising diffusion models in discrete state-spaces},
  author={Austin, Jacob and Johnson, Daniel D and Ho, Jonathan and Tarlow, Daniel and Van Den Berg, Rianne},
  journal={Advances in neural information processing systems},
  volume={34},
  pages={17981--17993},
  year={2021}
}

@article{xin2025lumina,
  title={Lumina-dimoo: An omni diffusion large language model for multi-modal generation and understanding},
  author={Xin, Yi and Qin, Qi and Luo, Siqi and Zhu, Kaiwen and Yan, Juncheng and Tai, Yan and Lei, Jiayi and Cao, Yuewen and Wang, Keqi and Wang, Yibin and others},
  journal={arXiv preprint arXiv:2510.06308},
  year={2025}
}
\bibliographystyle{iclr2026_conference}
\end{document}